\pdfoutput=1
\documentclass[10pt, logo, copyright]{nvidiatechreport}

\usepackage{mdframed}
\usepackage[utf8]{inputenc} 
\usepackage[T1]{fontenc}    

\usepackage{amsfonts}       
\usepackage{nicefrac}       
\usepackage{microtype}      
\usepackage[dvipsnames]{xcolor}         
\usepackage{multirow}
\usepackage{multicol}
\usepackage{graphicx}
\usepackage[numbers]{natbib}
\usepackage{tabto}
\usepackage{xspace}
\usepackage{amsmath}
\usepackage{adjustbox}
\usepackage{enumitem}
\usepackage{wrapfig}
\usepackage{dblfloatfix}

\usepackage{float}

\usepackage{natbib}

\usepackage{hyperref}
\usepackage{url}
\usepackage{booktabs}
\usepackage{multirow}
\usepackage{enumitem}
\usepackage{adjustbox}
\usepackage{tabularx}
\usepackage{arydshln}
\usepackage{wrapfig}

\usepackage{multirow}
\usepackage{colortbl}
\usepackage{amsmath}
\usepackage{makecell}
\usepackage{hhline}
\usepackage{array}
\usepackage{diagbox}
\usepackage{graphicx}
\usepackage{adjustbox}
\usepackage{subcaption}

\usepackage{listings}
\usepackage{minted}
\newcolumntype{H}{>{\setbox0=\hbox\bgroup}c<{\egroup}@{}}

\NewDocumentCommand{\mvk}
{ mO{} }{\textcolor{green}{\textsuperscript{\textit{matthijs}}\textsf{{\small[#1]}}}}

\NewDocumentCommand{\shizhe}
{ mO{} }{\textcolor{magenta}{\textsuperscript{\textit{shizhe}}\textsf{{\small[#1]}}}}

\NewDocumentCommand{\wb}
{ mO{} }{\textcolor{teal}{\textsuperscript{\textit{wonmin}}\textsf{{\small[#1]}}}}

\usepackage{fp} 

\usepackage{xcolor}
\definecolor{mygreen}{rgb}{0.7, 1.0, 0.7}

\definecolor{myblue}{rgb}{0.7, 0.8, 1.0}

\usepackage{pifont}
\newcommand\mynuma[1]{\ifcase#1 \or \ding{172}\or \ding{173}\or
  \ding{174}\or \ding{175}\or \ding{176}\or \ding{177}%
  \or \ding{178}\or \ding{179}\or \ding{180}\or \ding{181}\else *\fi\relax}

\newcommand\mynumb[1]{\ifcase#1 \or \ding{182}\or \ding{183}\or
  \ding{184}\or \ding{185}\or \ding{186}\or \ding{187}%
  \or \ding{188}\or \ding{189}\or \ding{190}\or \ding{191}\else *\fi\relax}

\title{\ours: Advancing Masked Discrete Diffusion for High-Resolution Image Synthesis}

\author{Shufan Li, Greg Heinrich, Hanrong Ye, Yonggan Fu, Aditya Grover, Jan Kautz, Pavlo Molchanov}

\correspondingauthor{X}

\begin{abstract}

 \textbf{Abstract:}  We propose \ours, a state-of-the-art masked discrete diffusion model (MDM) for high-resolution text-to-image synthesis. Compared with prior work on masked image generation, \ours~addresses two key challenges. First, unlike continuous diffusion models which progressively refine latent representations across the entire image, standard MDMs lack self-correcting capability because discrete tokens cannot be modified once they are unmasked. Second, although increasing the vocabulary size of discrete image tokenizers improves reconstruction fidelity, it introduces optimization difficulties for generative modeling as the per-token training signal becomes increasingly sparse. To address the first challenge, \ours~incorporates a token-editing mechanism that enables the model to dynamically revise already-unmasked tokens during inference, similar to how a sculptor iteratively refines their work. To tackle the second challenge, we propose a Grouped Cross-Entropy (GCE) objective that assigns positive learning signals to tokens neighboring the ground truth in embedding space, thereby alleviating signal sparsity. To further improve training efficiency, we implement a custom fused operator for GCE that significantly reduces VRAM usage in large-vocabulary settings. Experimental results demonstrate that these innovations substantially improve both training efficiency and image fidelity of masked discrete image generators, achieving a score of 0.90 on GenEval, 86.9 on DPG and  10.76 of HPSv3.
 
\end{abstract}

\newcommand{\greg}[1]{}
\newcommand{\PM}[1]{}

\newcommand{\ye}[1]{}

\newcommand{\ours}[0]{Nemotron-Labs-Diffusion-Image}

\begin{document}
\maketitle

    

\begin{figure*}[h]
    \centering
    \includegraphics[width=1\linewidth]{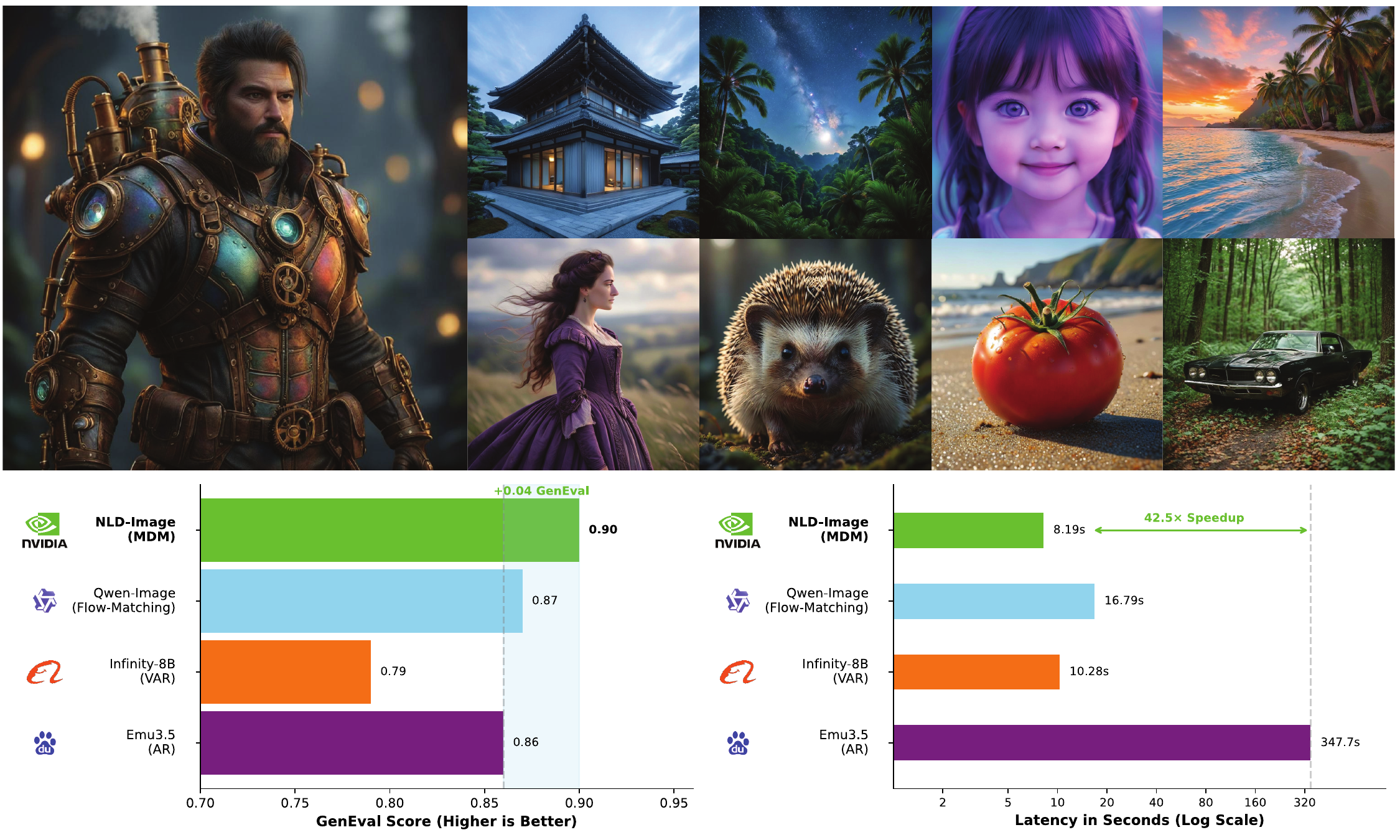}
    \caption{\textbf{We propose \ours(NLD-Image),} a masked discrete diffusion model for text-to-image synthesis. It achieves state-of-the-art performance at 1024px text-to-image synthesis, surpassing prior masked image generators. Images are sampled from prompts of the MJHQ dataset. }
    
    \label{fig:teaser}
\end{figure*}

\section{Introduction}

State-of-the-art text-to-image models \cite{openai2024gpt4o,flux2024,wu2025qwen,seedream2025seedream} achieve high-fidelity image synthesis primarily by scaling latent diffusion models (LDMs) \cite{rombach2022high} to large model sizes trained on massive text-image corpora. Compared with earlier alternatives such as GANs \cite{goodfellow2020generative}, the success of LDMs can largely be attributed to two key factors. First, VAEs provide a smooth and structured latent space that is easier to model than raw image pixels, making training more effective and scalable. Second, unlike GANs, which generate all pixels of an image in a single step, LDMs iteratively refine latent embeddings over multiple denoising steps, enabling progressive self-correction during inference. \PM{for people with LLM background, we need to put a table together showing different approaches to image generation: continuous, discrete, AR, MDM etc and highlight pros and cons, like quality, efficiency, training etc.}

Recently, there has been growing interest in discrete image generation, where images are first discretized into sequences of tokens and a generative model is trained to model the distribution over these token sequences \cite{bai2024meissonic,chang2022maskgit,wang2024emu3}. Compared with LDMs, discrete image generation offers several promising advantages. For example, discrete representations are naturally compatible with large language models (LLMs), which also operate on discrete tokens, making them an appealing foundation for unified multimodal models \cite{li2025lavidao,li2026lavida,yang2025mmada,shi2025muddit}. Furthermore, they can directly leverage well-established optimization techniques developed for LLM training and inference, such as sequence packing and pre-tokenization during training \cite{krell2021efficient}, as well as KV caching during inference, improving scalability and efficiency \cite{li2025sparse,ma2025dkv}.

Existing discrete image generators can be categorized into two families: autoregressive (AR) models and masked discrete diffusion models (MDMs). AR models generate image tokens sequentially following a raster-scan order, whereas MDMs start from a fully masked sequence and progressively unmask tokens over multiple diffusion steps. Importantly, MDMs can decode multiple tokens at arbitrary positions in parallel during each inference step, providing several advantages over AR models, including faster inference and native support for tasks such as image inpainting. As a result, recent state-of-the-art discrete image generators \cite{cui2025emu3,xie2024show} predominantly adopt the MDM paradigm.

In this work, we propose \ours, a state-of-the-art discrete image generator based on the masked diffusion paradigm. Compared with prior work, \ours~introduces several novel techniques that address two fundamental challenges in discrete image generation: (1) the lack of self-refinement capability and (2) the difficulty of training discrete image generators with large codebooks, which are essential for achieving high-fidelity image generation \cite{zhu2024scaling,chang2025scalable}.

The first challenge is the lack of self-refinement. Unlike LDMs, which iteratively refine outputs during inference, vanilla MDMs commit to each token once it is unmasked, preventing later correction even if the prediction is incorrect. This issue is particularly severe because tokens unmasked at each step are sampled independently from position-wise logits. Consequently, the effective joint distribution becomes a product of marginal token distributions, implicitly assuming inter-token independence. In practice, however, image tokens exhibit strong dependencies, creating a mismatch between the sampling process and the true data distribution, which leads to error accumulation in the final output.

To address this issue, we introduce a token editing mechanism that allows tokens to be corrected and refined even after they have been unmasked. We enable this capability by modifying the vanilla MDM training process with token corruption. Specifically, in addition to masking a subset of image tokens, \ours~also corrupts a portion of visible tokens and trains the model to predict token probabilities at all positions. For masked positions, the model predicts new tokens as in standard MDMs; for non-masked positions, it predicts whether existing tokens should be corrected. This design enables iterative refinement similar to continuous LDMs, reducing accumulated sampling errors and improving image fidelity.

\begin{figure*}[t]
    \centering
    \includegraphics[width=1\linewidth]{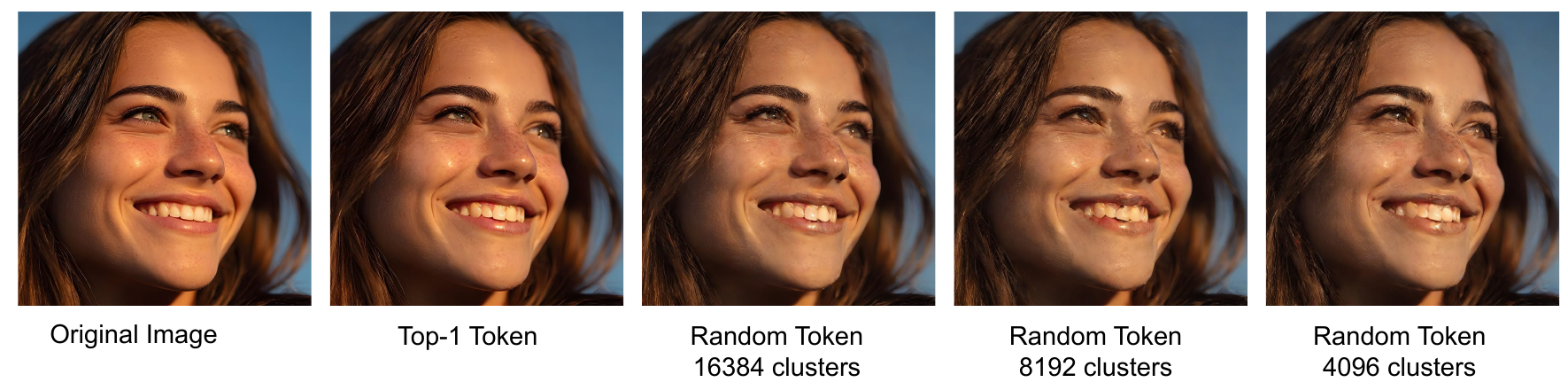}
    \caption{\textbf{Semantic similarities of adjacent tokens.} Using precomputed K-means clustering, we visualize the reconstruction results of the top-1 token and randomly selected tokens from the same cluster. The results show that these tokens are semantically similar, yet this relationship is not captured by the vanilla cross-entropy loss.}
    \label{fig:vis_cluster}
    \vspace{-10pt}
\end{figure*}

The second challenge concerns the difficulty of training generative models with large and expressive codebooks. Discrete image tokenizers typically use vector quantization to map continuous image features to discrete codes. Larger codebooks reduce discretization error and improve reconstruction fidelity \cite{shi2025scalable,zhu2024scaling}, but they also make downstream generative modeling substantially more difficult, often requiring larger model capacity and more training data \cite{cui2025emu3}. One key reason is the codebook sparsity problem \cite{li2026snce}: with a fixed image corpus, increasing the codebook size decreases the frequency of individual tokens, resulting in sparse and insufficient supervision signals. Two visually similar features that would map to the same code in a smaller codebook may instead map to different codes in a larger one. However, the standard one-hot cross-entropy objective fails to capture this semantic similarity and instead treats all non-ground-truth tokens as equally negative targets, creating optimization challenges. Figure~\ref{fig:vis_cluster} illustrates this phenomenon visually.


To mitigate this problem, we propose a grouped cross-entropy (GCE) objective that provides auxiliary supervision for non–top-1 tokens that are semantically close to the ground-truth token in embedding space. Given a large codebook (e.g., 100k entries), we first cluster the codes into a smaller set of groups (e.g., 16k) using K-means. During training, the model predicts probabilities over all tokens, and cluster-level probabilities are obtained by summing the probabilities of tokens within each cluster. We then apply a cross-entropy loss on the resulting cluster distribution using the ground-truth cluster label, alongside the standard token-level one-hot cross-entropy loss. We further extend this hierarchically with multiple clustering granularities, yielding multi-level supervision that encourages the model to capture semantic similarity beyond exact token matches. To improve efficiency, we implement a custom fused operator that significantly reduces the latency and VRAM overhead compared to a naive PyTorch autograd implementation.

We conduct extensive evaluations of \ours, and experimental results demonstrate strong performance across a wide range of text-to-image benchmarks, including GenEval~\cite{ghosh2023geneval}, DPG \cite{hu2024equipdpg}, and MJHQ-30k \cite{li2024playground}, highlighting the effectiveness of the proposed approaches. 

In summary, our contributions are as follows: 1) We propose \ours, an 8B state-of-the-art foundational masked discrete diffusion model for text-to-image synthesis, achieving strong performance across multiple text-to-image benchmarks. 2) We introduce a token editing mechanism that enables MDMs to iteratively refine and self-correct their outputs during inference, improving image fidelity. 3) We propose GCE, a novel training objective for large-vocabulary discrete image generators that alleviates the codebook sparsity problem and improves training efficiency and optimization stability.

\section{Background and Related Work}

\subsection{Discrete Image Generation}

Discrete image generators employ discrete image tokenizers \cite{zhu2024scaling,mentzer2023finite,yu2023language,shi2025scalable,chang2025scalable} to encode images into sequences of discrete codes and then learn to model these sequences. Early works primarily adopted autoregressive models \cite{van2017neural,ramesh2022hierarchical,yu2022scaling,sun2024autoregressive,chen2025janus,wang2024emu3}, which generate tokens sequentially. A key limitation of these models is their slow inference speed. To address this limitation, masked discrete diffusion models (MDMs) \cite{sahoo2024simple,shi2024simplified} learn to generate multiple tokens at each inference step, greatly reducing latency. MaskGIT~\cite{chang2022maskgit} pioneered masked generative image modeling. Meissonic~\cite{bai2024meissonic} further scaled this paradigm to high-resolution image synthesis through token compression. Several recent works have also explored unified understanding and generation models based on the discrete diffusion paradigm, including MMaDa~\cite{yang2025mmada}, the LaViDa-O series~\cite{li2025lavidao,li2025sparse,li2026lavida}, and Unidisc~\cite{hu2022unified}.

Concretely, MDMs begin inference from a fully masked sequence $y^1$ consisting entirely of the special mask token $[\text{M}]$. The model then gradually unmasks tokens over multiple sampling steps until a clean sequence $y^0$ is obtained. At intermediate timesteps $0 < t < 1$, the sequence $y^t$ contains a mixture of mask tokens and clean tokens. During training, given a clean sequence $y^0$, a random timestep $t$ is sampled uniformly from $[0,1]$, and a partially masked sequence $y^t$ is generated using the forward diffusion process $q(y^t|y^0)$, which randomly masks a subset of tokens in $y^0$. The model is then trained to predict the original tokens at masked positions using the following ELBO \PM{lets say what ELBO is} objective:

\begin{equation}
\mathcal{L}_{\text{ELBO}} =
\mathbb{E}_{y^0,\, t \sim \text{Unif}([0,1]),\, y^t \sim q(y^t|y^0)}
\left[
-\frac{1}{t} \sum_{i=1}^{L} 
\mathbf{I}\{y_i^t = [\text{M}]\}
\log p_\theta(y_i^0 \mid y^t)
\right]
\label{eq:mdm}
\end{equation}

\subsection{Token Editing}

During training, the objective in Equation~\ref{eq:mdm} computes the loss only on masked positions while treating non-mask tokens as correct. During inference, once a token is unmasked, it remains fixed, preventing MDMs from revising earlier decisions and leading to error accumulation. Recent works in the language domain have introduced self-correction mechanisms to address this limitation. Seed-Diffusion \cite{song2025seed} augments the forward process $q(y^t|y^0)$ with insertion and deletion operations and employs on-policy learning with Levenshtein-distance rewards. EditFlow \cite{havasi2025edit} extends MDM sampling with CMTC-based insertion and deletion operations. LLaDa-2.1 \cite{bie2026llada21} enables direct modification of previously unmasked tokens without explicit edit operations. However, extending token editing to discrete image generation remains largely unexplored. Image token sequences have fixed lengths, making insertion and deletion operations invalid, and it remains unclear which corruption strategies are effective for image synthesis. \ours~is the first work to introduce token-editing mechanisms for discrete image generation.

\subsection{Scaling Vocabulary Size}

Scaling codebook size is crucial for improving image fidelity in discrete image generation. Larger vocabularies make tokenizers more expressive, especially for large‑scale, high‑resolution text‑to‑image models \cite{cui2025emu3,team2026longcat}. However, large codebooks suffer from codebook sparsity: as vocabulary size grows, per‑token frequency drops sharply. With a dataset of 1M images and 256 tokens per image, an 8,192‑entry codebook yields an average frequency of 31,250 per code, but a 200K codebook reduces this to 1,280—only 4\% of the former. This sparsity makes training large‑vocabulary generators difficult. Existing work mainly relies on brute‑force scaling with >20B‑parameter models and >13T tokens of data \cite{cui2025emu3}. Concurrent work notes that discrete tokenizers underperform with limited data but improve as data scales \cite{team2026longcat}. SNCE \cite{li2026snce} mitigates sparsity via distance‑based soft labels, but gains are limited and soft labels increase memory cost. We propose a more efficient and effective GCE objective to address codebook sparsity.

\begin{figure*}
    \centering
    \includegraphics[width=1\linewidth]{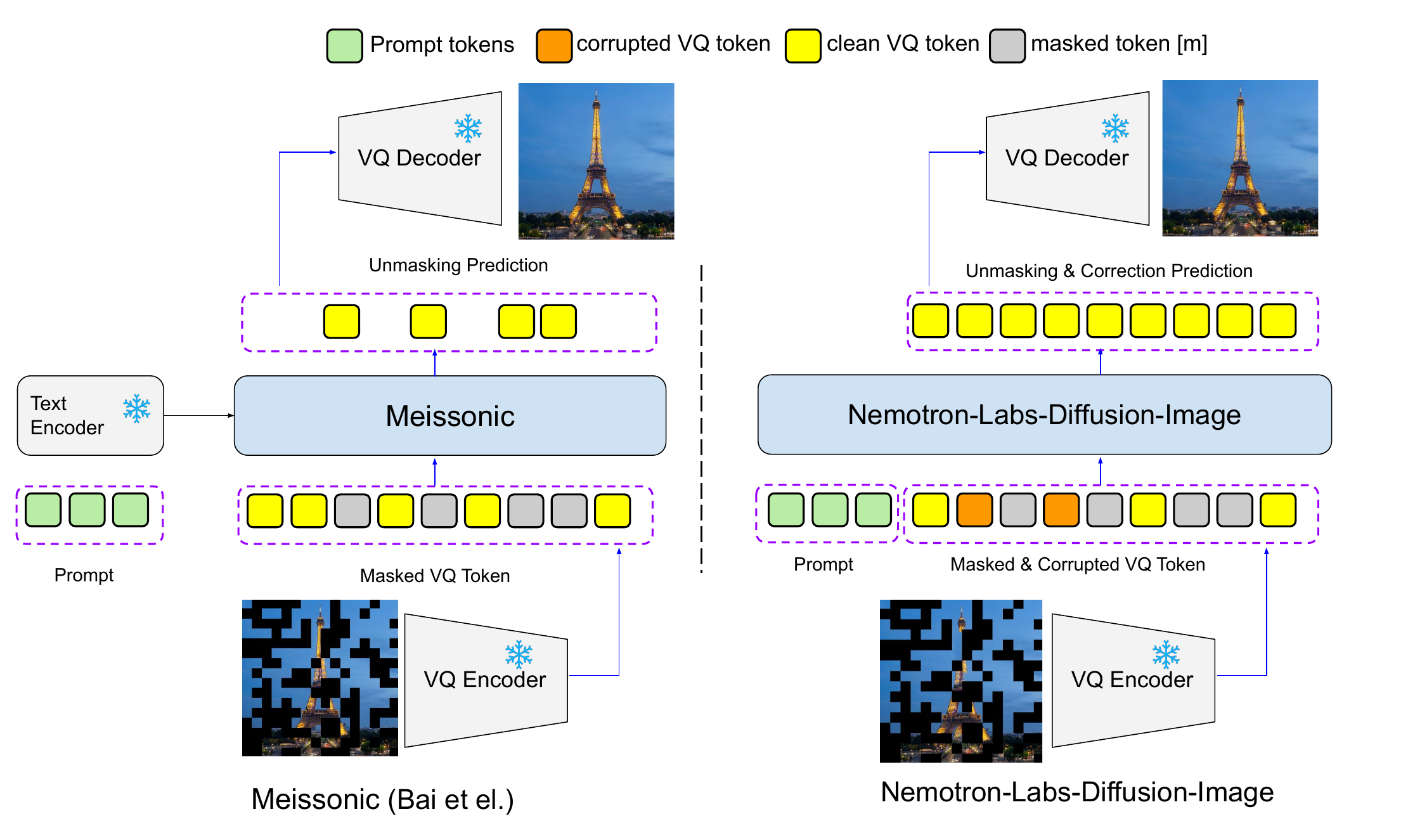}
    \caption{\textbf{Overall architecture of \ours.} Unlike prior works such as Meissonic, \ours~employs a single decoder-only transformer to process both text and image inputs. During training, it additionally corrupts clean image tokens to enable self-correction. *Corruption effects are amplified for better visual illustration. Figure \ref{fig:vis_cluster} provides a more accurate depiction. }
    \label{fig:architecture}
    \vspace{-20pt}
\end{figure*}

\section{Method}

\subsection{Model Architecture}

Prior works on discrete text-to-image models such as Meissonic \cite{bai2024meissonic} typically employ an encoder-decoder framework that combines a frozen text encoder with a trainable image generation network. By contrast, \ours~uses a single decoder-only transformer to process both text prompts and image tokens. This design is motivated by the recent success of unified understanding and generation models \cite{deng2025emerging,yang2025mmada}, which demonstrate that a single transformer can effectively model both text and images. In our experiments, we initialize our model from a pretrained diffusion language model \cite{fu2026nextron}, which is trained with masked modeling objectives on text tokens.

Compared with the encoder-decoder approach, \ours's decoder-only architecture offers several advantages. First, \ours~is not constrained by the context length of a frozen text encoder, allowing it to process substantially longer prompts. By contrast, Meissonic's CLIP encoder supports a maximum of only 77 tokens per prompt. Second, \ours~can naturally leverage sequence packing optimization \cite{krell2021efficient}, which is widely used in LLM pretraining to reduce unnecessary padding when prompts have varying lengths, thereby improving training efficiency. Third, by initializing \ours~from a pretrained diffusion language model, we can effectively utilize the language understanding capabilities of the base model and avoid learning text semantics from scratch, making training more efficient and stable. To adapt the base model for image generation, we replace the final language modeling head with a newly initialized MLP that predicts image tokens. We defer additional implementation details to Appendix~\ref{sec:appendix_results}. In total, \ours~contains 8B parameters optimized end-to-end.

\subsection{Self-Correction via Token Editing}

To equip \ours~with self-correction capabilities similar to those of continuous diffusion models, we introduce a token editing mechanism that enables the model to iteratively refine its outputs. Specifically, instead of predicting token probabilities only at masked positions, \ours~also predicts a ``correction distribution'' for clean image tokens.

\textbf{Inference.} At each inference step, given a partially unmasked sequence $y^t$, \ours~not only performs the standard unmasking process but also edits already unmasked tokens based on a confidence threshold. Specifically, if the model predicts a different token at an already unmasked position with confidence above a specified threshold $\tau$, we replace the original token with the newly predicted token. This process largely follows the setup of LLaDa-2.1 \cite{bie2026llada21}, a masked diffusion language model.

\textbf{Training.} To enable token editing capabilities, we modify the unmasking objective in Equation~\ref{eq:mdm} to incorporate a token-editing component. Specifically, given a clean sequence $y^0$ of length $L$, vanilla MDM training first samples a timestep $t \in [0,1]$ uniformly at random and then samples a partially masked sequence from the forward distribution $q(y^t|y^0)$. This is implemented by randomly replacing each clean image token with the special mask token $[\text{M}]$ with probability $p$. In expectation, $y^t$ contains $Lp$ masked tokens and $L(1-p)$ clean image tokens. \ours~extends this process by introducing additional corruption to the $L(1-p)$ clean tokens instead of directly copying them from the clean sequence $y^0$.

An important distinction from similar approaches in the language domain lies in the choice of corruption operations. In text generation, corruption often includes edit operations such as insertion and deletion \cite{ding2026beyond}. While these operations are natural for variable-length text generation, they are not applicable to image synthesis because the number of image tokens is fixed for a given resolution. Another common strategy is to replace clean tokens with random tokens uniformly sampled from the vocabulary \cite{zhang2025corrective}. However, we find this approach performs poorly in practice because randomly sampled tokens follow a substantially different distribution from incorrectly predicted tokens, leading to a mismatch between training and inference.

Instead, we employ two corruption strategies. In the first strategy, clean tokens are replaced with tokens randomly sampled from the same image. In the second strategy, clean tokens are replaced with random tokens selected from their nearest neighbors in the tokenizer embedding space. The overall training paradigm is illustrated in Figure~\ref{fig:architecture}. Formally, we denote this augmented forward process as $q'(y^t|y^0)$. The final objective is defined as follows:

\begin{equation}
\mathcal{L}_{\text{ELBO}_{\text{Edit}}} =
\mathbb{E}_{y^0,\, t \sim \text{Unif}([0,1]),\, y^t \sim q'(y^t|y^0)}
\left[
-\frac{1}{t} \sum_{i=1}^{L} 
\log p_\theta(y_i^0 \mid y^t)
\right]
\label{eq:mdm_edit}
\end{equation}

Compared with the vanilla MDM objective in Equation~\ref{eq:mdm}, we compute loss terms at all token positions rather than only masked positions. Additionally, $y^t$ now contains corrupted tokens in addition to clean and masked tokens. We defer further implementation details to Appendix~\ref{sec:appendix_edit}.

\begin{figure*}
    \centering
    \includegraphics[width=1\linewidth]{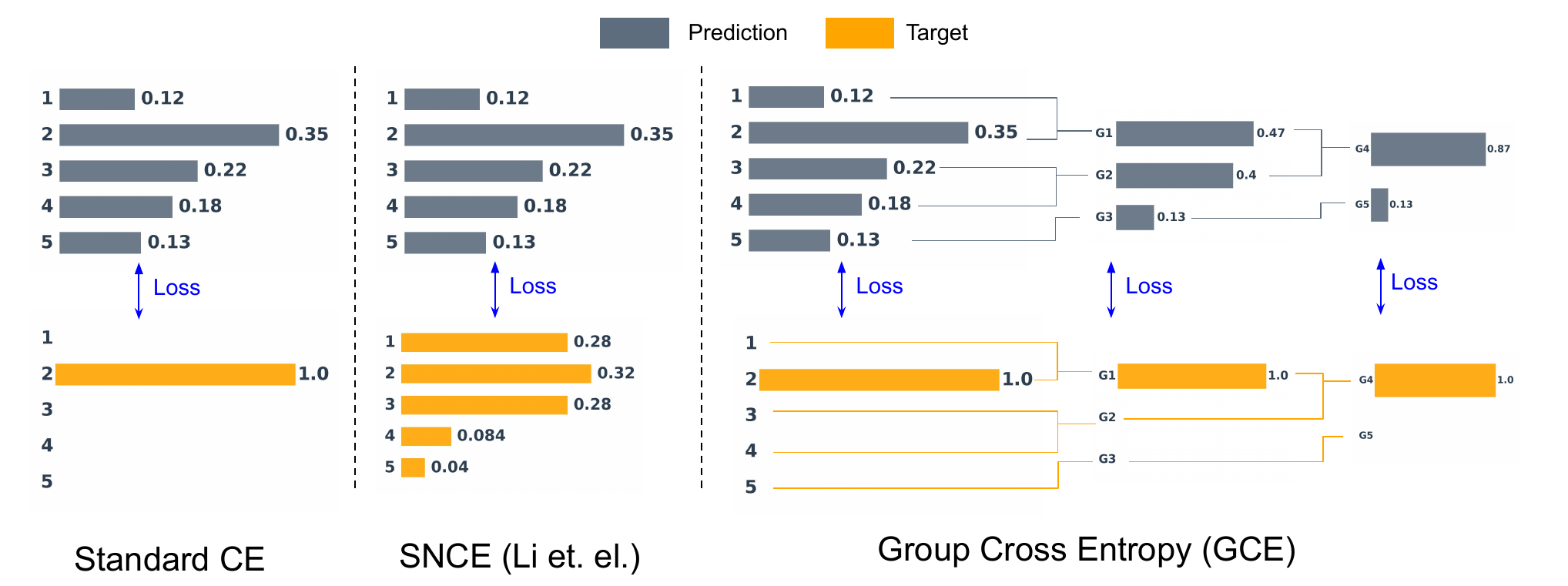}
    \caption{\textbf{Design of grouped cross-entropy (GCE).} (Left) Standard cross-entropy uses a one-hot target. (Middle) SNCE employs a fixed soft target. (Right) GCE groups codes into clusters and applies one-hot cross-entropy at multiple hierarchical levels.}
    \label{fig:gce_loss}
    \vspace{-15pt}
\end{figure*}

\subsection{Grouped Cross-Entropy Objective}

\label{sec:method_gce}

To address the sparse training signals caused by large codebooks, we design a hierarchical grouped cross-entropy objective that provides positive supervision for non-top-1 tokens that are semantically similar to the ground-truth token. Given a large codebook $V$ with size $|V|$, we first perform offline K-means clustering to group semantically similar tokens into $M$ clusters $C_1, C_2, \ldots, C_M$. During training, the model predicts unnormalized logits $h \in \mathbb{R}^{|V|}$. The probability of each code $i \in \{1,2,\ldots,|V|\}$ is obtained through the softmax operation $
p_i = \frac{\exp(h_i)}{\sum_{j=1}^{|V|} \exp(h_j)}.
$
We then derive cluster probabilities by summing the probabilities of individual tokens belonging to the same cluster:  $
\mathbb{P}(C_i) = \sum_{j \in C_i} p_j.
$

In Equation~\ref{eq:mdm_edit}, the term $\log p_\theta(y_i^0 \mid y^t)$ is implemented using a cross-entropy loss with one-hot targets. Similarly, we introduce auxiliary losses at the cluster level by applying cross-entropy losses with one-hot cluster labels on the cluster probabilities defined above. In practice, we use a tokenizer with a vocabulary size of 132K and include multiple clustering granularities with different numbers of clusters (e.g., 16,384 and 8,192). Let $C^j(\cdot)$ denote the cluster assignment under the $j$-th clustering. The final GCE objective is defined as:

\begin{equation}
J_{\text{GCE}}(y_i^0, y^t)
=
\underbrace{\log p_\theta(y_i^0 \mid y^t)}_{\text{top-1 term}}
+
\underbrace{
\sum_j \log \mathbb{P}\left(C^j(y_i^0) \mid y^t\right)
}_{\text{clustering terms}}
\label{eq:gce}
\end{equation}

During training, we replace $\log p_\theta(y_i^0 \mid y^t)$ in Equation~\ref{eq:mdm_edit} with $J_{\text{GCE}}$. For each clustering term in $J_{\text{GCE}}$, the gradient with respect to the unnormalized logits is given by:

\begin{equation}
\frac{\partial}{\partial h_i} \log P(C)
=
p_i
\left(
\frac{\mathbf{I}\{i \in C\}}{P(C)} - 1
\right)
\end{equation}

We make several observations. First, within each clustering term, all codes belonging to the target cluster $(i \in C)$ receive positive gradients, while all other codes receive negative gradients. Second, among tokens within the target cluster, the gradient magnitude is proportional to the post-softmax probability $p_i$, allowing the model to reinforce tokens that already have relatively high confidence. Third, when considering $J_{\text{GCE}}$ as a whole, tokens receive positive supervision proportional to their semantic proximity to the ground-truth token. For example, the ground-truth token receives positive signals from all terms, while semantically similar non-top-1 tokens that belong to the same fine-grained clusters receive positive signals from both fine-grained and coarse-grained clustering terms. Tokens that are farther away receive positive supervision from only a subset of clustering levels. These properties make $J_{\text{GCE}}$ particularly effective for mitigating the codebook sparsity problem. We visualize this design in Figure~\ref{fig:gce_loss}. The naive implementation of GCE based on PyTorch introduces considerable overhead. Hence, we designed a custom operator to accelerate GCE while reducing its memory overhead. We defer implementation details to Appendix \ref{sec:appendix_gce}.


\textbf{Comparison with SNCE.} SNCE \cite{li2026snce} replaces the one-hot target with distance-based soft labels, but GCE offers several advantages. First, SNCE uses a fixed smoothed target, whereas GCE allows the model to dynamically adjust its confidence. For example, if SNCE assigns a 0.7 soft target to the top-1 token but the model predicts 0.8, the top-1 token receives negative gradients, causing instability. In contrast, GCE always assigns positive gradients to the top-1 target, and its clustering terms can be optimized either by concentrating probability mass on the top-1 token or distributing it across tokens within the same cluster. Second, SNCE requires storing real-valued $L \times |V|$ soft targets, introducing additional memory overhead, while GCE stores only discrete cluster indices.  Third, SNCE optimizes a different objective: even with perfect predictions, its smoothed-label loss remains non-zero. In GCE, once the top-1 term is perfectly optimized, both the top-1 and clustering terms become zero. These properties make GCE a more efficient and effective alternative to SNCE.

\section{Experiments}

We initialize our model from an 8B pretrained diffusion language model \cite{fu2026nextron} and equip it with a tokenizer containing a 131K codebook \cite{cui2025emu3}. Following prior work \cite{bai2024meissonic,li2025lavidao}, we employ a progressive upscaling strategy that starts training at $256 \times 256$ resolution and gradually scales to $1024 \times 1024$ during training. We train the model for 300K steps on 64 H100 GPUs. Additional training details and dataset composition are deferred to Appendix~\ref{sec:appendix_results}. \PM{can we show training curves when initialized from scratch vs from pretrained weights?}

\subsection{Main Results on Large-Scale Text-to-Image Generation}

We report text-to-image generation results on the GenEval~\cite{ghosh2023geneval}, DPG~\cite{hu2024equipdpg}, and MJHQ~\cite{li2024playground} benchmarks. These results are presented in Table~\ref{tab:gen_eval_results} and Table~\ref{tab:dpg}. For GenEval and DPG, we report the corresponding benchmark scores. For MJHQ, we report both FID and HPSv3 metrics to evaluate image fidelity. We note that FID relies on an outdated feature extraction network trained on low-resolution images and therefore does not fully capture high-resolution image quality. Furthermore, multiple prior works have shown that FID correlates poorly with human perception of image quality \cite{podell2023sdxl,chen2025blip3}. By contrast, HPSv3 \cite{ma2025hpsv3} is a VLM-based reward model with a high-resolution image encoder specifically fine-tuned on human preference data, making it a more reliable indicator of image fidelity. We include FID primarily for completeness and consistency with prior literature.

We compare against specialist text-to-image models such as Flux-dev \cite{flux2024}, SD3-Medium \cite{esser2024scaling-sd3}, Meissonic \cite{bai2024meissonic}, and DALLE-3 \cite{openai_dalle3}, as well as unified multimodal models such as BAGEL \cite{deng2025emerging}, MMaDa \cite{yang2025mmada}, and LaViDa-O \cite{li2025lavidao}. Among models based on the masked diffusion paradigm, \ours~significantly outperforms the state-of-the-art specialist model Meissonic as well as unified multimodal models such as LaViDa-O. Notably, \ours~achieves competitive performance with frontier models such as Qwen-Image-2507 on GenEval and GPT-4o on the DPG benchmark, highlighting the effectiveness of our proposed approach.


\begin{table*}[t!]
  \centering
  \caption{\textbf{Text-to-Image Generation Performance on Geneval Benchmark.}}
  \label{tab:gen_eval_results}
  \begin{adjustbox}{max width=\textwidth}
  \begin{tabular}{ccccccccc}
    \toprule
    \textbf{Model} & \textbf{Params} & \textbf{Single Obj.$\uparrow$} & \textbf{Two Obj.$\uparrow$} & \textbf{Counting$\uparrow$} & \textbf{Colors$\uparrow$} & \textbf{Position$\uparrow$} & \textbf{Color Attri.$\uparrow$} & \textbf{Overall$\uparrow$} \\

    \midrule
\multicolumn{9}{c}{\textit{Unified MLLM}} \\

    Emu3~\cite{wang2024emu3} & 8B & - & - & - & - & - & - & 0.66  \\
    Janus-Pro~\cite{chen2025janus} & 7B & 0.99 & 0.89 & 0.59 & 0.90 & 0.79 & 0.66 & 0.80   \\
    MMaDA~\cite{yang2025mmada} & 8B & 0.99 & 0.76 & 0.61 & 0.84 & 0.20 & 0.37 & 0.63  \\
    Show-o~\cite{xie2024show} & 1.3B & 0.98 & 0.80 & 0.66 & 0.84 & 0.31 & 0.50 & 0.68 \\
    BAGEL~\cite{deng2025emerging} & 14B & 0.98 & 0.95 & 0.84 & 0.95 & 0.78 & 0.77 & 0.88 \\
    LaViDa-O~\cite{li2025lavidao} & 10B & 0.99 & 0.85 & 0.71 & 0.86 & 0.65 & 0.58 & 0.77  \\
    Show-o2~\cite{xie2025showo2} & 7B & 1.00 & 0.87 & 0.58 & 0.92 & 0.52 & 0.62 & 0.76  \\
        \midrule
  \multicolumn{9}{c}{\textit{Gen. Only}} \\
    PixArt-$\alpha$~\cite{chen2023pixartalpha} & 0.6B & 0.98 & 0.50 & 0.44 & 0.80 & 0.08 & 0.07 & 0.48 \\
    DALL-E 3~\cite{openai_dalle3} & - & 0.96 & 0.87 & 0.47 & 0.83 & 0.43 & 0.45 & 0.67 \\
    SD3-Medium~\cite{esser2024scaling-sd3} & 2B & 0.99 & 0.94 & 0.72 & 0.89 & 0.33 & 0.60 & 0.74  \\
    FLUX.1-dev~\cite{flux2024} & 12B & 0.98 & 0.81 & 0.74 & 0.79 & 0.22 & 0.45 & 0.66  \\

        Meissonic~\cite{bai2024meissonic} & 1B & 0.99 & 0.66 & 0.42 & 0.86 & 0.10 & 0.22 & 0.54  \\
    Qwen-Image-2507\cite{wu2025qwen} & 20B & 0.99 & 0.92&  0.89&  0.88&  0.76 & 0.77& 0.87  \\  
        \rowcolor{cyan!10}
    \ours & 8B &  0.98 & 0.93 & 0.83 & 0.94 & 0.88 & 0.82 & 0.90  \\
    
    \bottomrule
  \end{tabular}
  \end{adjustbox}
\end{table*}



\begin{table*}[t]
\centering
\caption{{\textbf{Text-to-Image Generation Performance on DPG Benchmark and MJHQ-30k Dataset.}} *Finetuned on 6M synthetic data for better image quality.}
\label{tab:dpg}
\setlength{\tabcolsep}{27pt}
\resizebox{1.00\linewidth}{!}{%
\begin{tabular}{lcccrc}

\toprule
\multirow{2}{*}{\textbf{Model}} &
\multirow{2}{*}{\textbf{Params}} &
\multirow{2}{*}{\textbf{Codebook}} &
\multirow{2}{*}{\textbf{DPG}$\uparrow$} &
\multicolumn{2}{c}{\textbf{MJHQ-30k}} \\
\cmidrule(lr){5-6}
 &  &  &  & \textbf{FID}$\downarrow$ & \textbf{HPSv3}$\uparrow$ \\

\midrule
SD3\cite{esser2024scaling-sd3} & 8B & - & 83.5 & 11.92 & 9.42 \\
GPT-4o \cite{esser2024scaling-sd3} & - & - & 85.3 & - & - \\
Flux-Dev\cite{flux2024} & 12B & - & - & 10.15 & -\\
\hline
Janus-Pro\cite{chen2025janus} & 7B & 16,384 & 84.1 & 10.10 & 8.81 \\ 
Emu3 \cite{wang2024emu3} & 8B & 32,678 & 80.6 & - & - \\
Show-o \cite{xie2024show} & 1B & 8,192 & - & 15.18 & 7.20 \\
MMaDa\cite{yang2025mmada} & 8B & 8,192 & 53.4 & 32.85 & 5.43 \\
LaViDa-O \cite{li2025lavidao} & 10B & 8,192 & 81.8 & 6.68 & 8.81 \\
\rowcolor{cyan!10}
\ours  & 8B & 131,072 & 85.2 & 6.46 & 9.61 \\
\rowcolor{cyan!10}
\ours*  & 8B & 131,072 & 86.9 & 12.23 & 10.76 \\
\bottomrule
\end{tabular}%
}
\end{table*}

\subsection{Ablations for Token Editing and Self-Correction}

\begin{figure}
    \centering
    \includegraphics[width=1.0\linewidth]{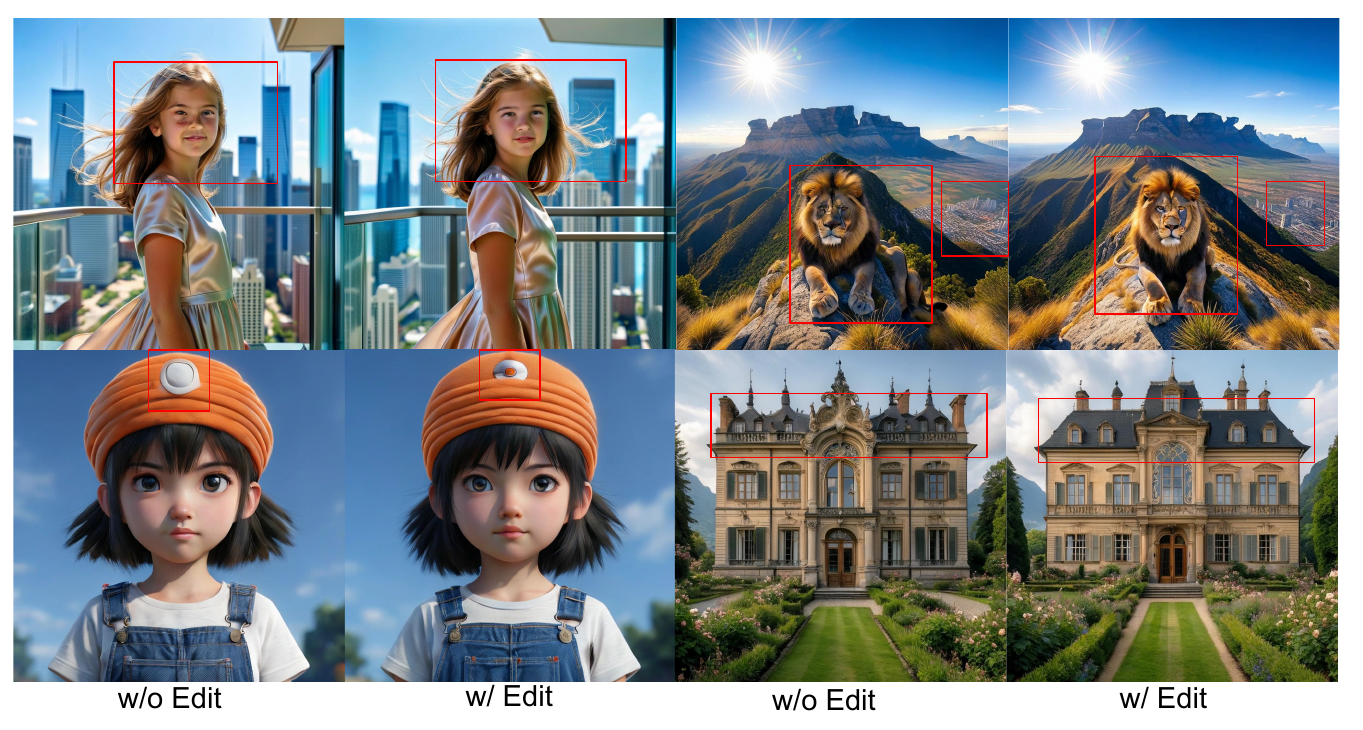}
 \caption{\textbf{Qualitative comparison of generated samples with and without token editing.}}
        \label{fig:vis_edit}
    \label{fig:placeholder}
\end{figure}

To validate the effectiveness of our proposed token editing pipeline and investigate whether self-correction improves image fidelity, we conduct both qualitative and quantitative evaluations on the MJHQ dataset. In Figure~\ref{fig:vis_edit}, we fix the random seed and visually compare images generated with and without the token editing pipeline. Token editing consistently improves image fidelity by correcting artifacts and refining texture details in the generated images.

\begin{figure*}[t]
    \centering

    \begin{subfigure}{0.60\linewidth}
        \centering
        \includegraphics[width=\linewidth]{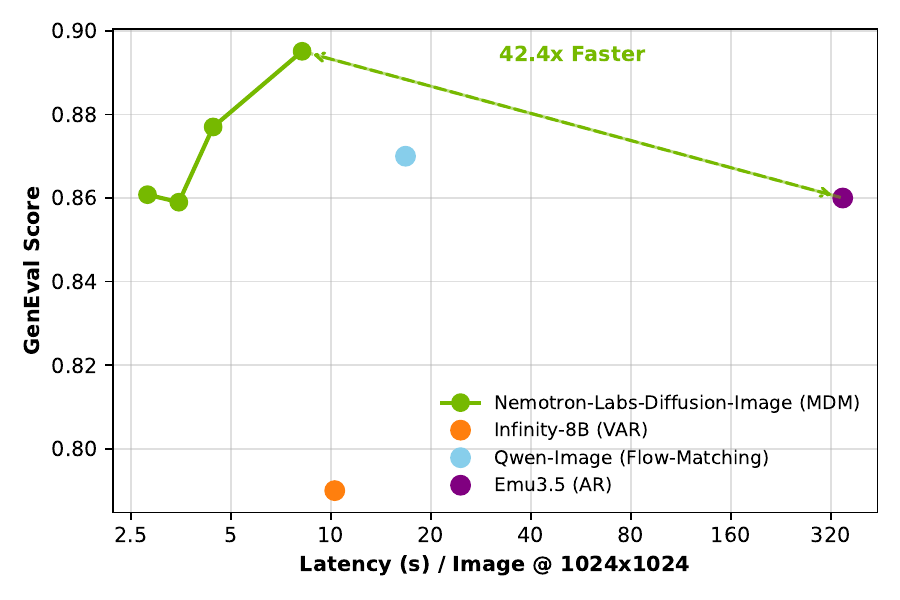}
    \caption{Comparision of GenEval score and inference latency between \ours~ and other  state-of-the-art models. Results are measured with a batch size of 1 on a H100 GPU. }
        \label{fig:latency_speedup}
    \end{subfigure}
    \hfill
    \begin{subfigure}{0.34\linewidth}
        \centering
        \includegraphics[width=\linewidth]{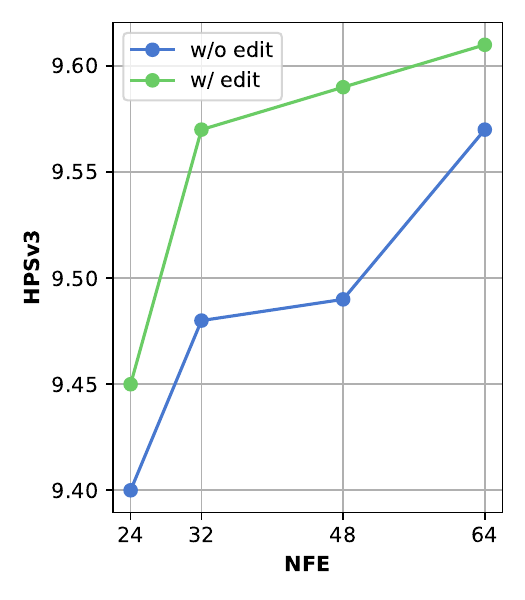}
        \caption{HPSv3 scores under different numbers of sampling steps with and without token editing. }
        \label{fig:nfe_hps}
    \end{subfigure}

    \caption{\textbf{Speed-Quality Tradeoff of  \ours. }}
    \label{fig:placeholder_combined}
\end{figure*}

In Figure~\ref{fig:nfe_hps}, we report HPSv3 scores on the MJHQ dataset with and without token editing. We evaluate \ours~under different numbers of sampling steps, also known as the Number of Function Evaluations (NFEs). We draw two main conclusions from these results. First, token editing consistently improves image quality across all NFEs. Second, although reducing NFEs decreases image quality in both settings, the degradation is substantially smoother when token editing is enabled. Notably, generations with token editing at 32 NFEs achieve performance comparable to generations without token editing at 64 NFEs, representing an effective 2$\times$ reduction in forward calls given a fixed target quality.

\subsection{Ablations for the Grouped Cross-Entropy Objective}

To fairly evaluate the effectiveness of our proposed GCE objective and demonstrate its advantages over alternatives such as SNCE, we conduct extensive ablation studies in a controlled setting. Following the setup of SNCE \cite{li2026snce}, we perform experiments on class-conditional image generation using the ImageNet~\cite{russakovsky2015imagenet} dataset at $256 \times 256$ resolution. We use the exact same model architecture, tokenizer, and training schedule as SNCE \cite{li2026snce}, with the optimization objective being the only varying factor. Results are reported in Table~\ref{tab:snce_results}. GCE consistently outperforms both SNCE and the vanilla cross-entropy baseline, demonstrating its effectiveness and advantages.

\begin{table*}[ht]
\centering
\caption{{\textbf{Class-conditioned Image Synthesis on ImageNet256 dataset.}}  *Models have identical-sized transformer layers. Parameter count increased due to larger token embedding and final linear head.}
 \label{tab:snce_results}
  \begin{adjustbox}{max width=\textwidth}
  \setlength{\tabcolsep}{13pt}
\begin{tabular}{lcccccc}
\toprule
\textbf{Objective} & \textbf{Params} & \textbf{Tokenizer} &\textbf{Tokenizer Pretraining} &\textbf{Codebook} & \textbf{Epoch} & \textbf{FID} $\downarrow$ \\
\midrule
CE  & 577M* &  Emu3.5-IBQ \cite{cui2025emu3} & Large-Scale T2I & 131,072 & 100 & 7.53 \\
SNCE & 577M* & Emu3.5-IBQ \cite{cui2025emu3}  & Large-Scale T2I  &131,072 & 100 & 3.62 \\
        \rowcolor{cyan!10}
GCE & 577M* & Emu3.5-IBQ \cite{cui2025emu3}  & Large-Scale T2I  &131,072 & 100 & 3.40 \\
\midrule
CE &  577M*  & Emu3.5-IBQ \cite{cui2025emu3} & Large-Scale T2I &131,072 & 300 & 5.44 \\
SNCE & 577M* & Emu3.5-IBQ \cite{cui2025emu3} & Large-Scale T2I & 131,072 & 300 & 3.42\\
        \rowcolor{cyan!10}
GCE & 577M* & Emu3.5-IBQ \cite{cui2025emu3}  & Large-Scale T2I  &131,072 & 300 & 3.00 \\

\midrule
CE & 846M* & FVQ \cite{zhu2024scaling} & ImageNet256 & 262,144 & 300 & 4.11 \\
SNCE &  846M*& FVQ \cite{zhu2024scaling} &ImageNet256 & 262,144 & 300 & 
3.20 \\
        \rowcolor{cyan!10}
GCE &  846M*& FVQ \cite{zhu2024scaling} &ImageNet256 & 262,144 & 300 & 
2.69 \\

\bottomrule
\end{tabular}
\end{adjustbox}
\end{table*}

\begin{figure}[ht!]
    \centering
    \includegraphics[width=1\linewidth]{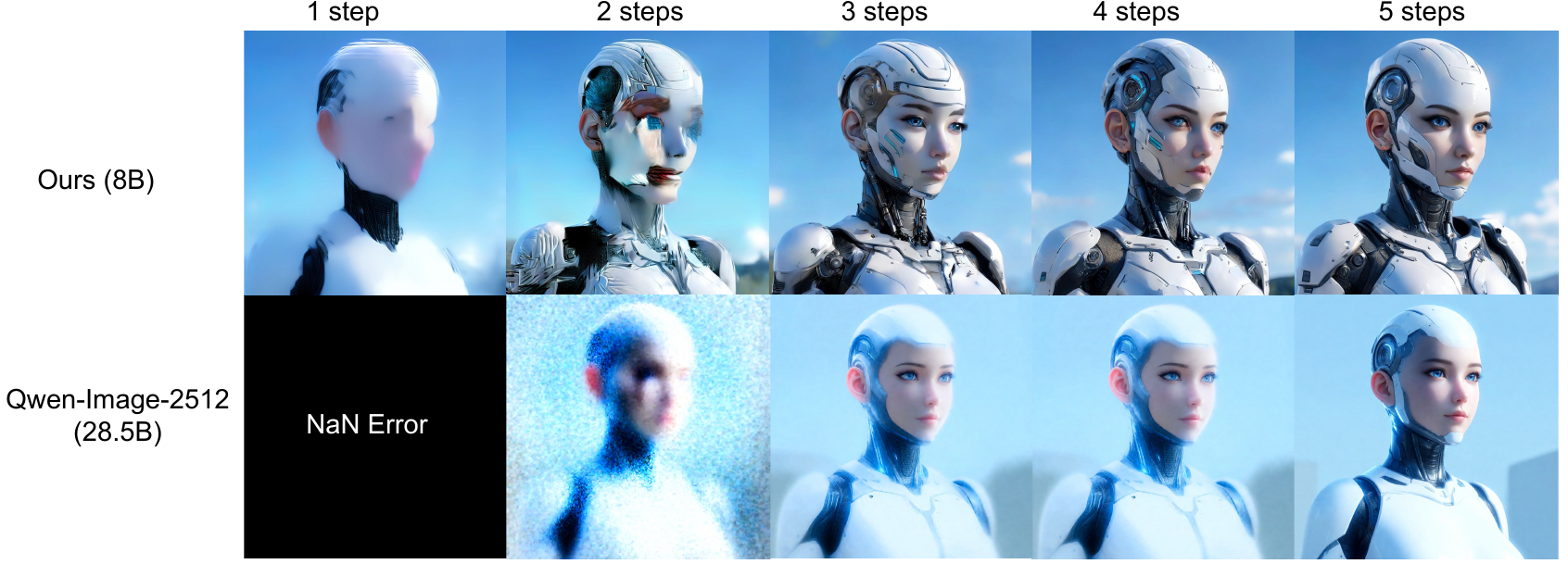}
    \caption{\textbf{Few-Step Generation Results.} We visualize the sampled images with 1,2,3,4,5 total steps and compare with a continuous model Qwen-Image \cite{wu2025qwen}.}
    \label{fig:progression}
\end{figure}

\begin{figure}[ht!]
    \centering
    \includegraphics[width=1\linewidth]{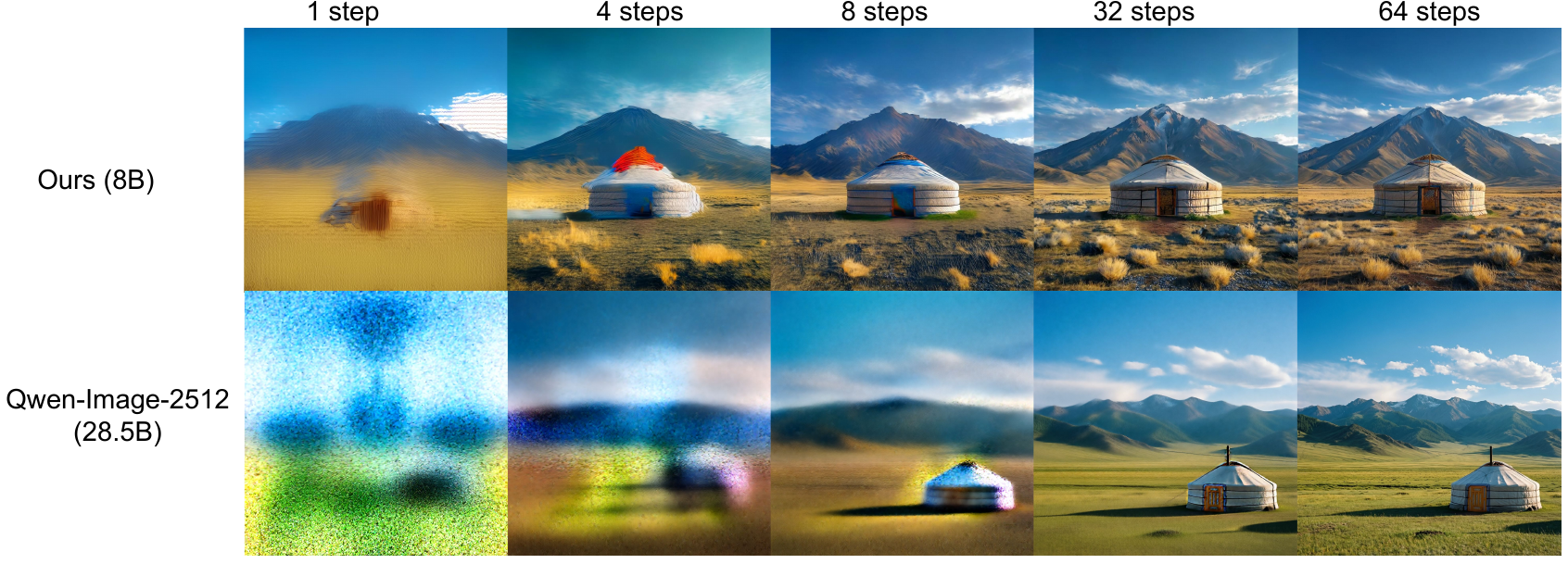}
    \caption{\textbf{Denoising Dynamics throughout the sampling process.} We visualize the $y^0$ prediction at 1st,4th,8th,32th, and 64th step out of a total of 64 steps.}
    \label{fig:progression_dynamic}
\end{figure}

\begin{table}[ht!]
\centering
\caption{\textbf{Performance Comparison of Loss Function Implementations.}}
\label{tab:loss-performance}
  \begin{adjustbox}{max width=\textwidth}
\begin{tabular}{lcccc}
\toprule
\textbf{Operation} & \textbf{Latency $\downarrow$} & \textbf{Input VRAM$\downarrow$} & \textbf{Active VRAM$\downarrow$} & \textbf{Max VRAM$\downarrow$} \\
\midrule
\texttt{F.cross\_entropy} (hard label) & 12.71 ms & 8.1 GB & 8.1 GB & 16.1 GB \\
\texttt{F.cross\_entropy} (soft label) & 25.00 ms & 16.1 GB & 8.1 GB & 24.2 GB \\
        \rowcolor{cyan!10}
GCE (eager) & 44.14 ms & 8.1 GB & 17.1 GB & 25.2 GB \\
        \rowcolor{cyan!10}
GCE (optimized forward) & 17.86 ms & 8.1 GB & 16.1 GB & 24.2 GB \\
        \rowcolor{cyan!10}
GCE (optimized fwd \& bwd) & 20.04 ms & 8.1 GB & 8.1 GB & 16.1 GB \\
\bottomrule
\end{tabular}
\end{adjustbox}
\end{table}

\subsection{Compute Efficiency}

\textbf{Optimized Operator.} To validate the effectiveness of our optimized operator, we benchmark both VRAM usage and latency when processing 16,384 tokens on a H100 GPU, corresponding to either 4 images at $1024 \times 1024$ resolution or 64 images at $256 \times 256$ resolution. We report latency, input tensor VRAM usage, and maximum VRAM consumption during both forward and backward computation. We define active VRAM as the difference between maximum VRAM and input VRAM, which measures the additional memory overhead introduced by loss computation. For fairness, we enable $\texttt{torch.compile}$ in all experiments.  We report results in Table \ref{tab:loss-performance} Compared with the eager implementation, our optimized operator reduces latency from 44\,ms to 20\,ms while decreasing maximum VRAM usage from 25\,GB to 16\,GB. 

Compared with the standard cross-entropy baseline using one-hot labels, our method introduces no additional memory overhead and only an 8\,ms increase in latency. We note that this overhead is negligible given that the overall training time is approximately 3.2\,s per step. All measurements are conducted on H100 GPUs.

\textbf{Generation Latency.} We also compare the inference latency of \ours~against state-of-the-art text=to-image models of including Qwen-Image (flow-matching), Infinity-8B (VAR) and Emu3.5 (AR). These results are visualized in Figure \ref{fig:latency_speedup}. Compared with Emu3.5, \ours~is 42.4$\times$ faster while also acheveing higher GenEval score.

\textbf{Few-Step Generation.} In most experiments, we use 64 diffusion steps to generate image samples. We also explored few-step generation and show qualitative results in Figure \ref{fig:progression}. Compared with continuous flow-matching models which predicts a blurry mean field at few-step setting, \ours~can generate images with reasonable quality at only 4 steps. This capability naturally emerges without any distillation process, highlighting another advantage of discrete diffusion models.

\textbf{Denoising Dynamics.} To further understand why \ours~works well with few sampling steps out of box,  we visualize the denoising dynamics at different stages of the 64-step sampling schedule in Figure \ref{fig:progression_dynamic}. We observe that the clean image prediction $y^0$ for \ours~converges faster to reasonable quality, while the clean image prediction for the continuous baseline Qwen-Image \cite{wu2025qwen} remain noisy and blurry. This illustration highlight the effectiveness of the discrete diffusion process in image synthesis.


\section{Conclusion}

We introduced \ours, a new state-of-the-art masked discrete diffusion model for high-resolution text-to-image synthesis. To overcome the inherent limitations of standard masked discrete models, \ours~incorporates a novel token-editing mechanism that enables iterative refinement during inference. In addition, we proposed the Grouped Cross-Entropy (GCE) objective, which alleviates the sparsity of training signals in large-vocabulary discrete token spaces by assigning positive supervision to semantically neighboring tokens in the embedding space. To further enhance practical scalability, we designed a custom fused operator for GCE that significantly reduces memory footprint and computational overhead, enabling efficient scaling. Experiments demonstrate that \ours~achieves substantial improvements in both image fidelity and training efficiency compared to prior masked image generation methods, paving the way for more powerful and scalable discrete image generation.

{
  \small
  \bibliographystyle{unsrt}
  \bibliography{main}

@article{openai2024gpt4o,
  title={GPT-4o System Card},
  author={OpenAI},
  journal={arXiv preprint arXiv:2410.21276},
  year={2024},
  url={https://arxiv.org/abs/2410.21276}
}

@article{lou2023discrete-sedd,
  title={Discrete diffusion modeling by estimating the ratios of the data distribution},
  author={Lou, Aaron and Meng, Chenlin and Ermon, Stefano},
  journal={arXiv preprint arXiv:2310.16834},
  year={2023}
}

@article{sahoo2024simple,
  title={Simple and effective masked diffusion language models},
  author={Sahoo, Subham and Arriola, Marianne and Schiff, Yair and Gokaslan, Aaron and Marroquin, Edgar and Chiu, Justin and Rush, Alexander and Kuleshov, Volodymyr},
  journal={Advances in Neural Information Processing Systems},
  volume={37},
  pages={130136--130184},
  year={2024}
}

@inproceedings{esser2024scaling-sd3,
  title={Scaling rectified flow transformers for high-resolution image synthesis},
  author={Esser, Patrick and Kulal, Sumith and Blattmann, Andreas and Entezari, Rahim and M{\"u}ller, Jonas and Saini, Harry and Levi, Yam and Lorenz, Dominik and Sauer, Axel and Boesel, Frederic and others},
  booktitle={Forty-first international conference on machine learning},
  year={2024}
}

@inproceedings{rombach2022high,
  title={High-resolution image synthesis with latent diffusion models},
  author={Rombach, Robin and Blattmann, Andreas and Lorenz, Dominik and Esser, Patrick and Ommer, Bj{\"o}rn},
  booktitle={Proceedings of the IEEE/CVF conference on computer vision and pattern recognition},
  pages={10684--10695},
  year={2022}
}

@article{podell2023sdxl,
  title={Sdxl: Improving latent diffusion models for high-resolution image synthesis},
  author={Podell, Dustin and English, Zion and Lacey, Kyle and Blattmann, Andreas and Dockhorn, Tim and M{\"u}ller, Jonas and Penna, Joe and Rombach, Robin},
  journal={arXiv preprint arXiv:2307.01952},
  year={2023}
}

@inproceedings{chang2022maskgit,
  title={Maskgit: Masked generative image transformer},
  author={Chang, Huiwen and Zhang, Han and Jiang, Lu and Liu, Ce and Freeman, William T},
  booktitle={Proceedings of the IEEE/CVF conference on computer vision and pattern recognition},
  pages={11315--11325},
  year={2022}
}

@inproceedings{radford2021learning,
  title={Learning transferable visual models from natural language supervision},
  author={Radford, Alec and Kim, Jong Wook and Hallacy, Chris and Ramesh, Aditya and Goh, Gabriel and Agarwal, Sandhini and Sastry, Girish and Askell, Amanda and Mishkin, Pamela and Clark, Jack and others},
  booktitle={International conference on machine learning},
  pages={8748--8763},
  year={2021},
  organization={PmLR}
}

@article{hu2022unified,
  title = {Unified Discrete Diffusion for Simultaneous Vision-Language Generation},
  author = {Hu, Minghui and Zheng, Chuanxia and Zheng, Heliang and Cham, Tat-Jen and Wang, Chaoyue and Yang, Zuopeng and Tao, Dacheng and Suganthan, Ponnuthurai N},
  journal = {arXiv},
  year = {2022},
}

@article{yang2025mmada,
  title   = {Multimodal Large Diffusion Language Models},
  author  = {Yang, Ling and Tian, Ye and Li, Bowen and Zhang, Xinchen and Shen, Ke and Tong, Yunhai and Wang, Mengdi},
  journal = {arXiv preprint arXiv:2505.15809},
  year    = {2025}
}

@article{you2025lladav,
  title={Llada-v: Large language diffusion models with visual instruction tuning},
  author={You, Zebin and Nie, Shen and Zhang, Xiaolu and Hu, Jun and Zhou, Jun and Lu, Zhiwu and Wen, Ji-Rong and Li, Chongxuan},
  journal={arXiv preprint arXiv:2505.16933},
  year={2025}
}

@article{li2025lavida,
  title={Lavida: A large diffusion language model for multimodal understanding},
  author={Li, Shufan and Kallidromitis, Konstantinos and Bansal, Hritik and Gokul, Akash and Kato, Yusuke and Kozuka, Kazuki and Kuen, Jason and Lin, Zhe and Chang, Kai-Wei and Grover, Aditya},
  journal={arXiv preprint arXiv:2505.16839},
  year={2025}
}

@article{chen2025blip3,
  title={Blip3-o: A family of fully open unified multimodal models-architecture, training and dataset},
  author={Chen, Jiuhai and Xu, Zhiyang and Pan, Xichen and Hu, Yushi and Qin, Can and Goldstein, Tom and Huang, Lifu and Zhou, Tianyi and Xie, Saining and Savarese, Silvio and others},
  journal={arXiv preprint arXiv:2505.09568},
  year={2025}
}

@article{deng2025emerging,
  title={Emerging properties in unified multimodal pretraining},
  author={Deng, Chaorui and Zhu, Deyao and Li, Kunchang and Gou, Chenhui and Li, Feng and Wang, Zeyu and Zhong, Shu and Yu, Weihao and Nie, Xiaonan and Song, Ziang and others},
  journal={arXiv preprint arXiv:2505.14683},
  year={2025}
}

@article{chen2025janus,
  title={Janus-pro: Unified multimodal understanding and generation with data and model scaling},
  author={Chen, Xiaokang and Wu, Zhiyu and Liu, Xingchao and Pan, Zizheng and Liu, Wen and Xie, Zhenda and Yu, Xingkai and Ruan, Chong},
  journal={arXiv preprint arXiv:2501.17811},
  year={2025}
}

@article{bai2024meissonic,
  title={Meissonic: Revitalizing masked generative transformers for efficient high-resolution text-to-image synthesis},
  author={Bai, Jinbin and Ye, Tian and Chow, Wei and Song, Enxin and Li, Xiangtai and Dong, Zhen and Zhu, Lei and Yan, Shuicheng},
  journal={arXiv preprint arXiv:2410.08261},
  year={2024}
}

@article{shi2025muddit,
  title={Muddit: Liberating generation beyond text-to-image with a unified discrete diffusion model},
  author={Shi, Qingyu and Bai, Jinbin and Zhao, Zhuoran and Chai, Wenhao and Yu, Kaidong and Wu, Jianzong and Song, Shuangyong and Tong, Yunhai and Li, Xiangtai and Li, Xuelong and others},
  journal={arXiv preprint arXiv:2505.23606},
  year={2025}
}

@misc{flux2024,
    author={Black Forest Labs},
    title={FLUX},
    year={2024},
    howpublished={\url{https://github.com/black-forest-labs/flux}},
}

@article{ghosh2023geneval,
  title={Geneval: An object-focused framework for evaluating text-to-image alignment},
  author={Ghosh, Dhruba and Hajishirzi, Hannaneh and Schmidt, Ludwig},
  journal={Advances in Neural Information Processing Systems},
  volume={36},
  pages={52132--52152},
  year={2023}
}

@article{hu2024equipdpg,
  title={Equip diffusion models with llm for enhanced semantic alignment},
  author={Hu, Xiwei and Wang, Rui and Fang, Yixiao and Fu, Bin and Cheng, Pei and Ella, Gang Yu},
  journal={arXiv preprint arXiv:2403.05135},
  volume={5},
  number={7},
  pages={16},
  year={2024}
}

@article{xie2024show,
  title={Show-o: One single transformer to unify multimodal understanding and generation},
  author={Xie, Jinheng and Mao, Weijia and Bai, Zechen and Zhang, David Junhao and Wang, Weihao and Lin, Kevin Qinghong and Gu, Yuchao and Chen, Zhijie and Yang, Zhenheng and Shou, Mike Zheng},
  journal={arXiv preprint arXiv:2408.12528},
  year={2024}
}

@article{xie2025showo2,
  title={Show-o2: Improved Native Unified Multimodal Models},
  author={Xie, Jinheng and Yang, Zhenheng and Shou, Mike Zheng},
  journal={arXiv preprint arXiv:2506.15564},
  year={2025}
}

@misc{openai_dalle3,
  author       = {OpenAI},
  title        = {DALL·E 3},
  year         = {2023},
  howpublished = {\url{https://openai.com/index/dall-e-3/}},
}

@misc{li2024playground,
      title={Playground v2.5: Three Insights towards Enhancing Aesthetic Quality in Text-to-Image Generation}, 
      author={Daiqing Li and Aleks Kamko and Ehsan Akhgari and Ali Sabet and Linmiao Xu and Suhail Doshi},
      year={2024},
      eprint={2402.17245},
      archivePrefix={arXiv},
      primaryClass={cs.CV}
}

@misc{laion-aesthetics,
title={LAION-AESTHETICS},
howpublished={\url{https://laion.ai/blog/laion-aesthetics/}},
note={Accessed: 2024 - 03 - 06},
author = {Christoph Schuhmann},
year = {2022}
}

@article{schuhmann2022laion,
  title={Laion-5b: An open large-scale dataset for training next generation image-text models},
  author={Schuhmann, Christoph and Beaumont, Romain and Vencu, Richard and Gordon, Cade and Wightman, Ross and Cherti, Mehdi and Coombes, Theo and Katta, Aarush and Mullis, Clayton and Wortsman, Mitchell and others},
  journal={Advances in neural information processing systems},
  volume={35},
  pages={25278--25294},
  year={2022}
}

@misc{kakaobrain2022coyo-700m,
  title         = {COYO-700M: Image-Text Pair Dataset},
  author        = {Byeon, Minwoo and Park, Beomhee and Kim, Haecheon and Lee, Sungjun and Baek, Woonhyuk and Kim, Saehoon},
  year          = {2022},
  howpublished  = {\url{https://github.com/kakaobrain/coyo-dataset}},
}

@article{chen2025sharegpt,
  title={ShareGPT-4o-Image: Aligning Multimodal Models with GPT-4o-Level Image Generation},
  author={Chen, Junying and Cai, Zhenyang and Chen, Pengcheng and Chen, Shunian and Ji, Ke and Wang, Xidong and Yang, Yunjin and Wang, Benyou},
  journal={arXiv preprint arXiv:2506.18095},
  year={2025}
}

@article{li2025lavidao,
  title={Lavida-O: Elastic Masked Diffusion Models for Unified Multimodal Understanding and Generation},
  author={Li, Shufan and Gu, Jiuxiang and Liu, Kangning and Lin, Zhe and Wei, Zijun and Grover, Aditya and Kuen, Jason},
  journal={arXiv preprint arXiv:2509.19244},
  year={2025}
}

@article{ma2025dkv,
  title={dkv-cache: The cache for diffusion language models},
  author={Ma, Xinyin and Yu, Runpeng and Fang, Gongfan and Wang, Xinchao},
  journal={arXiv preprint arXiv:2505.15781},
  year={2025}
}

@inproceedings{ma2025hpsv3,
  title={Hpsv3: Towards wide-spectrum human preference score},
  author={Ma, Yuhang and Wu, Xiaoshi and Sun, Keqiang and Li, Hongsheng},
  booktitle={Proceedings of the IEEE/CVF International Conference on Computer Vision},
  pages={15086--15095},
  year={2025}
}

@article{van2017neural,
  title={Neural discrete representation learning},
  author={Van Den Oord, Aaron and Vinyals, Oriol and others},
  journal={Advances in neural information processing systems},
  volume={30},
  year={2017}
}

@article{mentzer2023finite,
  title={Finite scalar quantization: Vq-vae made simple},
  author={Mentzer, Fabian and Minnen, David and Agustsson, Eirikur and Tschannen, Michael},
  journal={arXiv preprint arXiv:2309.15505},
  year={2023}
}

@article{yu2023language,
  title={Language Model Beats Diffusion--Tokenizer is Key to Visual Generation},
  author={Yu, Lijun and Lezama, Jos{\'e} and Gundavarapu, Nitesh B and Versari, Luca and Sohn, Kihyuk and Minnen, David and Cheng, Yong and Birodkar, Vighnesh and Gupta, Agrim and Gu, Xiuye and others},
  journal={arXiv preprint arXiv:2310.05737},
  year={2023}
}

@inproceedings{shi2025scalable,
  title={Scalable image tokenization with index backpropagation quantization},
  author={Shi, Fengyuan and Luo, Zhuoyan and Ge, Yixiao and Yang, Yujiu and Shan, Ying and Wang, Limin},
  booktitle={Proceedings of the IEEE/CVF International Conference on Computer Vision},
  pages={16037--16046},
  year={2025}
}

@article{zhu2024scaling,
  title={Scaling the codebook size of vq-gan to 100,000 with a utilization rate of 99\%},
  author={Zhu, Lei and Wei, Fangyun and Lu, Yanye and Chen, Dong},
  journal={Advances in Neural Information Processing Systems},
  volume={37},
  pages={12612--12635},
  year={2024}
}

@article{yu2022scaling,
  title={Scaling autoregressive models for content-rich text-to-image generation},
  author={Yu, Jiahui and Xu, Yuanzhong and Koh, Jing Yu and Luong, Thang and Baid, Gunjan and Wang, Zirui and Vasudevan, Vijay and Ku, Alexander and Yang, Yinfei and Ayan, Burcu Karagol and others},
  journal={arXiv preprint arXiv:2206.10789},
  volume={2},
  number={3},
  pages={5},
  year={2022}
}

@article{sun2024autoregressive,
  title={Autoregressive model beats diffusion: Llama for scalable image generation},
  author={Sun, Peize and Jiang, Yi and Chen, Shoufa and Zhang, Shilong and Peng, Bingyue and Luo, Ping and Yuan, Zehuan},
  journal={arXiv preprint arXiv:2406.06525},
  year={2024}
}

@article{wang2024emu3,
  title={Emu3: Next-token prediction is all you need},
  author={Wang, Xinlong and Zhang, Xiaosong and Luo, Zhengxiong and Sun, Quan and Cui, Yufeng and Wang, Jinsheng and Zhang, Fan and Wang, Yueze and Li, Zhen and Yu, Qiying and others},
  journal={arXiv preprint arXiv:2409.18869},
  year={2024}
}

@article{li2025sparse,
  title={Sparse-LaViDa: Sparse Multimodal Discrete Diffusion Language Models},
  author={Li, Shufan and Gu, Jiuxiang and Liu, Kangning and Lin, Zhe and Wei, Zijun and Grover, Aditya and Kuen, Jason},
  journal={arXiv preprint arXiv:2512.14008},
  year={2025}
}

@article{li2026lavida,
  title={LaViDa-R1: Advancing Reasoning for Unified Multimodal Diffusion Language Models},
  author={Li, Shufan and Zhu, Yuchen and Gu, Jiuxiang and Liu, Kangning and Lin, Zhe and Chen, Yongxin and Tao, Molei and Grover, Aditya and Kuen, Jason},
  journal={arXiv preprint arXiv:2602.14147},
  year={2026}
}

@article{cui2025emu3,
  title={Emu3. 5: Native multimodal models are world learners},
  author={Cui, Yufeng and Chen, Honghao and Deng, Haoge and Huang, Xu and Li, Xinghang and Liu, Jirong and Liu, Yang and Luo, Zhuoyan and Wang, Jinsheng and Wang, Wenxuan and others},
  journal={arXiv preprint arXiv:2510.26583},
  year={2025}
}

@article{russakovsky2015imagenet,
  title={Imagenet large scale visual recognition challenge},
  author={Russakovsky, Olga and Deng, Jia and Su, Hao and Krause, Jonathan and Satheesh, Sanjeev and Ma, Sean and Huang, Zhiheng and Karpathy, Andrej and Khosla, Aditya and Bernstein, Michael and others},
  journal={International journal of computer vision},
  volume={115},
  number={3},
  pages={211--252},
  year={2015},
  publisher={Springer}
}

@article{chang2025scalable,
  title={Scalable training for vector-quantized networks with 100\% codebook utilization},
  author={Chang, Yifan and Qin, Jie and Qiao, Limeng and Wang, Xiaofeng and Zhu, Zheng and Ma, Lin and Wang, Xingang},
  journal={arXiv preprint arXiv:2509.10140},
  year={2025}
}

@misc{chen2023pixartalpha,
      title={PixArt-$\alpha$: Fast Training of Diffusion Transformer for Photorealistic Text-to-Image Synthesis}, 
      author={Junsong Chen and Jincheng Yu and Chongjian Ge and Lewei Yao and Enze Xie and Yue Wu and Zhongdao Wang and James Kwok and Ping Luo and Huchuan Lu and Zhenguo Li},
      year={2023},
      eprint={2310.00426},
      archivePrefix={arXiv},
      primaryClass={cs.CV}
}

@article{wu2025qwen,
  title={Qwen-image technical report},
  author={Wu, Chenfei and Li, Jiahao and Zhou, Jingren and Lin, Junyang and Gao, Kaiyuan and Yan, Kun and Yin, Sheng-ming and Bai, Shuai and Xu, Xiao and Chen, Yilei and others},
  journal={arXiv preprint arXiv:2508.02324},
  year={2025}
}

@article{seedream2025seedream,
  title={Seedream 4.0: Toward next-generation multimodal image generation},
  author={Seedream, Team and Chen, Yunpeng and Gao, Yu and Gong, Lixue and Guo, Meng and Guo, Qiushan and Guo, Zhiyao and Hou, Xiaoxia and Huang, Weilin and Huang, Yixuan and others},
  journal={arXiv preprint arXiv:2509.20427},
  year={2025}
}

@article{goodfellow2020generative,
  title={Generative adversarial networks},
  author={Goodfellow, Ian and Pouget-Abadie, Jean and Mirza, Mehdi and Xu, Bing and Warde-Farley, David and Ozair, Sherjil and Courville, Aaron and Bengio, Yoshua},
  journal={Communications of the ACM},
  volume={63},
  number={11},
  pages={139--144},
  year={2020},
  publisher={ACM New York, NY, USA}
}

@article{krell2021efficient,
  title={Efficient sequence packing without cross-contamination: Accelerating large language models without impacting performance},
  author={Krell, Mario Michael and Kosec, Matej and Perez, Sergio P and Fitzgibbon, Andrew},
  journal={arXiv preprint arXiv:2107.02027},
  year={2021}
}

@article{li2026snce,
  title={SNCE: Geometry-Aware Supervision for Scalable Discrete Image Generation},
  author={Li, Shufan and Gu, Jiuxiang and Liu, Kangning and Lin, Zhe and Grover, Aditya and Kuen, Jason},
  journal={arXiv preprint arXiv:2603.15150},
  year={2026}
}

@article{ramesh2022hierarchical,
  title={Hierarchical text-conditional image generation with clip latents},
  author={Ramesh, Aditya and Dhariwal, Prafulla and Nichol, Alex and Chu, Casey and Chen, Mark},
  journal={arXiv preprint arXiv:2204.06125},
  volume={1},
  number={2},
  pages={3},
  year={2022}
}

@article{song2025seed,
  title={Seed diffusion: A large-scale diffusion language model with high-speed inference},
  author={Song, Yuxuan and Zhang, Zheng and Luo, Cheng and Gao, Pengyang and Xia, Fan and Luo, Hao and Li, Zheng and Yang, Yuehang and Yu, Hongli and Qu, Xingwei and others},
  journal={arXiv preprint arXiv:2508.02193},
  year={2025}
}

@article{havasi2025edit,
  title={Edit Flows: Flow Matching with Edit Operations},
  author={Havasi, Marton and Karrer, Brian and Gat, Itai and Chen, Ricky TQ},
  journal={arXiv preprint arXiv:2506.09018},
  year={2025}
}

@article{bie2026llada21,
  title={Llada2. 1: Speeding up text diffusion via token editing},
  author={Bie, Tiwei and Cao, Maosong and Cao, Xiang and Chen, Bingsen and Chen, Fuyuan and Chen, Kun and Du, Lun and Feng, Daozhuo and Feng, Haibo and Gong, Mingliang and others},
  journal={arXiv preprint arXiv:2602.08676},
  year={2026}
}

@article{team2026longcat,
  title={Longcat-next: Lexicalizing modalities as discrete tokens},
  author={Team, Meituan LongCat and Xiao, Bin and Wang, Chao and Li, Chengjiang and Zhang, Chi and Peng, Chong and Yu, Hang and Yang, Hao and Yan, Haonan and Sun, Haoze and others},
  journal={arXiv preprint arXiv:2603.27538},
  year={2026}
}

@article{ding2026beyond,
  title={Beyond Masks: Efficient, Flexible Diffusion Language Models via Deletion-Insertion Processes},
  author={Ding, Fangyu and Ding, Ding and Chen, Sijin and Wang, Kaibo and Xu, Peng and Feng, Zijin and Bai, Haoli and Han, Kai and Yan, Youliang and Yuan, Binhang and others},
  journal={arXiv preprint arXiv:2603.23507},
  year={2026}
}

@article{zhang2025corrective,
  title={Corrective Diffusion Language Models},
  author={Zhang, Shuibai and Peng, Fred Zhangzhi and Zhang, Yiheng and Pan, Jin and Chrysos, Grigorios G},
  journal={arXiv preprint arXiv:2512.15596},
  year={2025}
}

@article{bai2025qwen3,
  title={Qwen3-vl technical report},
  author={Bai, Shuai and Cai, Yuxuan and Chen, Ruizhe and Chen, Keqin and Chen, Xionghui and Cheng, Zesen and Deng, Lianghao and Ding, Wei and Gao, Chang and Ge, Chunjiang and others},
  journal={arXiv preprint arXiv:2511.21631},
  year={2025}
}

@article{fu2026nextron,
title = {Nemotron-Labs-Diffusion: A Tri-Mode Language Model Unifying Autoregressive, Diffusion, and Self-Speculation Decoding},
author = {Fu, Yonggan and Whalen, Lexington and Garg, Abhinav and Wu, Chengyue and Khadkevich, Maksim and Oswald, Nicolai and Xie, Enze and Egert, Daniel and Sreenivas, Sharath Turuvekere and Diao, Shizhe and Yu, Chenhan and Yu, Ye and Chen, Weijia and Norouzi, Sajad and Lan, Shiyi and Zhu, Ligeng and Wang, Jin and Jiang, Jindong and Mardani, Morteza and Maghoumi, Mehran and Han, Song and Jukic, Ante and Tajbakhsh, Nima and Kautz, Jan and Molchanov, Pavlo},
journal = {preprint},
year = {2026},
month = may,
}

@article{shi2024simplified,
  title={Simplified and generalized masked diffusion for discrete data},
  author={Shi, Jiaxin and Han, Kehang and Wang, Zhe and Doucet, Arnaud and Titsias, Michalis},
  journal={Advances in neural information processing systems},
  volume={37},
  pages={103131--103167},
  year={2024}
}
}

\appendix
\section{Additional Technical Details}

\subsection{Formulation of Masked Diffusion Models}

In this section, we provide an overview of the standard formulation of discrete diffusion models that are adopted by the literature \cite{sahoo2024simple,lou2023discrete-sedd,you2025lladav,li2025lavida,li2025lavidao}. Notations are adapted from these references to be consistent with the ones used in the main paper to avoid potential confusion. 

Given a sequence $y^0$ consisting of $L$ discrete tokens $y^0_1...y^0_L$, the forward discrete diffusion process $q(y^t|y^s)$ gradually replace the original tokens in $y^0$ to a special mask token [M] over the time interval $[0,1]$, with $1 \ge t \ge s \ge 0$. At $t=1$, the sequence $y^1$ is a fully masked sequence. This forward process is formally defined as

\newcommand{\cat}[0]{\text{Cat}}
\newcommand{\alphats}[0]{\frac{1-t}{1-s}}
\newcommand{\oneminusalphats}[0]{\frac{t-s}{1-s}}


\begin{equation}
    q(y^t_i|y^s_i) =  
    \begin{cases}
      \cat(y^t_i;\textbf{M}), & \text{if } y^s_i=[M] \\
      \cat(y^t_i;\alphats \mathbf{Y}^s_i+\oneminusalphats \textbf{M}), & \text{if } y^s_i \ne [M],
    \end{cases}
\end{equation}

where $\cat(\cdot)$ denotes a discrete categorical distribution.  $\textbf{M}, \mathbf{Y^s_i} \in \mathbb{R}^{|V|}$ are one-hot probability vectors, and $|V|$ is the vocabulary size. Specifically, $\textbf{M}$ is the one-hot vector of the special token $[M]$, and $\mathbf{y^s_i}$ is a one-hot vector of the token $y^s_i$. It has been shown that this forward process has the following marginal:

\begin{equation}
    q(y^t_i|y^0_i) =  \cat(y^t_i;(1-t) \mathbf{Y}^0_i+t \textbf{M}).
    \label{eq:q_process}
\end{equation}

MDLM \cite{sahoo2024simple} shows that the posterior of the reverse process $p(y^s|y^t,y^0)$ has the following form:

\begin{equation}
    p(y^s_i|y^t_i,Y^0_i) =  
    \begin{cases}
      \cat(y^s_i;\mathbf{y}^t_i), & \text{if } y^s_i \ne [M] \\
      \cat(y^s_i;\tfrac{t-s}{t} \mathbf{Y}^0_i+\tfrac{s}{t} \textbf{M}), & \text{if } y^s_i = [M].
    \end{cases}
    \label{eq:appendix-eq-p}
\end{equation}

At inference, the clean sequence $y^0$ is not known at start, so it is substituted with the prediction from a masked diffusion model $p_\theta(Y^0_i|y^t)$, which gives the following empirical sampling process:

\begin{equation}
    p_\theta(y^s_i|y^t) =  
    \begin{cases}
      \cat(y^s_i;\mathbf{y}^t_i), & \text{if } y^s_i \ne [M] \\
      \cat(y^s_i;\tfrac{t-s}{t} p_\theta(Y^0_i|y^t)+\tfrac{s}{t} \textbf{M}), & \text{if } X_s^i = [M].
    \end{cases}
    \label{eq:appendix-inference}
\end{equation}




When sampling a sequence from $p_\theta$, we initialize $y^1$ as a fully masked sequence and iterative applies equation \ref{eq:appendix-inference} until we reach $y^0$. Notably, once token $i$ is unmasked at timestep $t$, then $y_i^s\ne[M]$ holds  for all timesteps $s$ such that $s<t$, and we have $p_\theta(y^s_i|y^t)=\cat(y^s_i;\mathbf{y}^t_i)$.  This means that it will not change in subsequent sampling steps. 

During training, we optimize the maximum likelihood objective

\begin{equation}
\mathcal{L}_{\text{MDM}}=-\mathbb{E}_{(y,x)\sim\mathcal{D}} [\log p_\theta(y|x)]
    \label{eq:mle}
\end{equation}

However, $\log p_\theta(y|x)$ is intractable and requires integrating over all possible trajectories, we instead optimize the ELBO described in equation \ref{eq:mdm} from the main paper:

\begin{equation}
\mathcal{L}_{\text{ELBO}} =
\mathbb{E}_{y^0,\, t \sim \text{Unif}([0,1]),\, y^t \sim q(y^t|y^0)}
\left[
-\frac{1}{t} \sum_{i=1}^{L} 
\mathbf{I}\{y_i^t = [\text{M}]\}
\log p_\theta(y_i^0 \mid y^t)
\right]
\label{eq:mdm_appendix_loss}
\end{equation}

We can safely introduce the indicator term $\mathbf{I}\{y_i^t = [\text{M}]\}$ because when $y_i^t = [\text{M}]$, $p_\theta(y^s_i|y^t)=\cat(y^s_i;\mathbf{y}^t_i)$ does not depend on $p_\theta$. However, these assumptions no longer holds when token editing is introduced.
\subsection{Token Editing}
\label{sec:appendix_edit}

In this section, we provide a detailed description of the token editing process.

\textbf{Training.} To enable token editing, we modify the forward process $q(y_i^t|y_i^0)$ to $q'(y_i^t|y_i^0)$ by introducing a corruption term:

\begin{equation}
    q'(y_i^t|y_i^0) =
    \cat(y_i^t;
    (1-t)(1-\alpha)\mathbf{Y}_i^0
    + (1-t)\alpha \mathbf{C}(y_i^0)
    + t\mathbf{M}).
    \label{eq:q_process_modified}
\end{equation}

where $\alpha$ denotes the corruption level and $\mathbf{C}:V\rightarrow\mathbb{R}^{|V|}$ defines the distribution of corrupted tokens conditioned on the original clean token $y_i^0$.

In our implementation, we design $\mathbf{C}$ such that its probability mass is distributed over tokens from the same image as well as neighboring tokens in the embedding space. To obtain neighboring tokens, we directly reuse the precomputed K-means clustering used for the GCE objective. As shown in Figure~\ref{fig:vis_cluster} of the main paper, tokens from the same cluster typically produce visually similar images with only minor degradation in quality, making them a good approximation of erroneous model predictions during inference.

Using the augmented forward process $q'(y_i^t|y_i^0)$, we optimize the edit-aware ELBO objective defined in Equation~\ref{eq:mdm_edit} of the main paper.

\textbf{Inference.} We largely adopt the inference pipeline from LLaDa-2.1 \cite{bie2026llada21}, which explores token editing for text diffusion models. Specifically, at each inference step, we first perform the standard unmasking operation defined in Equation~\ref{eq:appendix-inference}. In addition, we compute $p_\theta(y_i^0|y^t)$ at non-masked positions. We edit token $y_i^t$ whenever the model predicts an alternative token $\widehat{y}_i^t$ with confidence above a predefined threshold $\tau$.

In our experiments, we find that $\tau=0.6$ and $\alpha=0.1$ work best. Detailed ablation results are provided in Appendix~\ref{sec:appendix_results}.

\subsection{Grouped Cross-Entropy}

\label{sec:appendix_gce}

In this section, we provide a detailed description of GCE. Recall from Section~\ref{sec:method_gce} of the main paper that each clustering term is defined as

\begin{align}
    \log \mathbb{P}(C(y_i^0)|y^t)
    =
    \log \left(
    \sum_{j\in C(y_i^0)} p_j
    \right).
\end{align}

We can further express this term using the unnormalized logits $h_j$:

\begin{align}
   \log \left(
   \sum_{j\in C(y_i^0)} p_j
   \right)
   &=
   \log \left(
   \sum_{j\in C(y_i^0)}
   \frac{\exp(h_j)}
   {\sum_{k=1}^{|V|}\exp(h_k)}
   \right) \\
   &=
   \log \left(
   \frac{
   \sum_{j\in C(y_i^0)}\exp(h_j)
   }{
   \sum_{k=1}^{|V|}\exp(h_k)
   }
   \right) \\
   &=
   \log \left(
   \sum_{j\in C(y_i^0)}\exp(h_j)
   \right)
   -
   \log \left(
   \sum_{k=1}^{|V|}\exp(h_k)
   \right).
\end{align}

Given a batch of logits represented as a tensor of shape $N\times |V|$, where $N=\text{SeqLen}\times\text{NumSeqs}$, the second term can be efficiently implemented using the optimized \texttt{torch.logsumexp} operator over the vocabulary dimension. The first term, however, is more challenging because clusters have different sizes.

A naive implementation would mask logits by setting codes outside the cluster $C(y_i^0)$ to $-\infty$. However, this approach requires allocating a full copy of the logits tensor, leading to substantial memory and latency overhead. To address this issue, we note that cluster sizes are bounded. Instead of allocating a tensor of size $N\times |V|$, we only allocate a tensor of size $N\times |C_{\max}|$, where $|C_{\max}|$ denotes the size of the largest cluster. Empirically, this value is 391 when using 8,192 clusters and 192 when using 16,384 clusters. In both cases, this requires less than 1\% of the memory needed by the naive implementation.

During the backward pass, instead of relying on autograd, we manually compute gradients using

\begin{equation}
    \frac{\partial}{\partial h_i}\log P(C)
    =
    p_i
    \left(
    \frac{\mathbf{I}\{i\in C\}}{P(C)} - 1
    \right)
    =
    p_i\frac{\mathbf{I}\{i\in C\}}{P(C)}
    - p_i.
\end{equation}

Although $p_i$ is dense, the first term is non-zero only for tokens within the target cluster. Therefore, we can apply a similar optimization strategy and allocate only $N\times |C_{\max}|$ memory for the sparse term, then combine the two terms efficiently using in-place \texttt{torch.scatter\_add} operations. We provide the full PyTorch implementation below:
\definecolor{codegreen}{rgb}{0,0.6,0}
\definecolor{codegray}{rgb}{0.5,0.5,0.5}
\definecolor{codepurple}{rgb}{0.58,0,0.82}
\definecolor{backcolour}{rgb}{0.95,0.95,0.92}

\lstdefinestyle{mystyle}{
    backgroundcolor=\color{backcolour},   
    commentstyle=\color{codegreen},
    keywordstyle=\color{magenta},
    numberstyle=\tiny\color{codegray},
    stringstyle=\color{codepurple},
    basicstyle=\ttfamily\footnotesize,
    breakatwhitespace=false,         
    breaklines=true,                 
    captionpos=b,                    
    keepspaces=true,                 
    numbers=left,                    
    numbersep=5pt,                  
    showspaces=false,                
    showstringspaces=false,
    showtabs=false,                  
    tabsize=2
}
\lstset{style=mystyle}

\begin{lstlisting}[language=Python]

class OptimalCappedGroupedCE(torch.autograd.Function):
    @staticmethod
    def forward(ctx, logits, target_clusters, cluster_map, cluster_sizes, cap):
        logits_f32 = logits.float()
        
        # log_z = log sum(exp(h_i)) for all i in V
        log_z = torch.logsumexp(logits_f32, dim=1, keepdim=True)

        # Map targets to their cluster members
        relevant_indices = cluster_map[target_clusters]
        relevant_logits = torch.gather(logits_f32, 1, relevant_indices)

        # Apply capacity mask to the cluster
        mask = torch.arange(cap, device=logits.device).unsqueeze(0) < \
               cluster_sizes[target_clusters].unsqueeze(1)
        masked_relevant = torch.where(mask, relevant_logits, torch.tensor(-1e20))
        
        # log_num = log sum(exp(h_k)) for k in C_y
        log_num = torch.logsumexp(masked_relevant, dim=1, keepdim=True)

        loss = (log_z - log_num).mean()

        ctx.save_for_backward(logits_f32, log_z, log_num, relevant_indices, mask)
        ctx.original_dtype = logits.dtype
        return loss

    @staticmethod
    def backward(ctx, grad_output):
        logits_f32, log_z, log_num, relevant_indices, mask = ctx.saved_tensors
        
        # Gradient Step 1: Compute p_i for all classes (Dense)
        grad_logits = torch.exp(logits_f32 - log_z)
        
        # Gradient Step 2: Compute pi / P(C_y) for cluster classes (Sparse)
        relevant_logits = torch.gather(logits_f32, 1, relevant_indices)
        target_grads = torch.exp(relevant_logits - log_num) * mask.float()
        
        # Gradient Step 3: In-place subtraction (Memory Efficient)
        # grad = p_i - pi / P(C_y)
        grad_logits.scatter_add_(1, relevant_indices, -target_grads)
        
        # Scale by external gradient and batch size
        grad_logits.mul_(grad_output / logits_f32.shape[0])
        
        return grad_logits.to(ctx.original_dtype), None, None, None, None
\end{lstlisting}

Note $J_{GCE}$ is a maximization objective. To be compatible with semantic of vanilla Pytorch cross-entropy loss which is a minimization objective, the loss term and gradients in the code are negated. 
\section{Additional Experiment Details and Results}
\label{sec:appendix_results}

In this section, we provide additional training details and experimental results, including qualitative samples and ablation studies.

\subsection{Training Data}

Our training dataset consists of 137M text-image pairs sourced from public datasets. The data pipeline largely follows the prior work LaViDa-O \cite{li2025lavidao}. Specifically, we source raw images from LAION-2B~\cite{schuhmann2022laion}, COYO-700M~\cite{kakaobrain2022coyo-700m}, BLIP3o-60k~\cite{chen2025blip3}, and ShareGPT4o-Image~\cite{chen2025sharegpt}. These datasets are heavily filtered to remove NSFW prompts, low CLIP-score samples~\cite{radford2021learning}, low aesthetic-score samples~\cite{laion-aesthetics}, and low-resolution images.

For all images from LAION-2B and COYO-700M, we use Qwen3-VL \cite{bai2025qwen3} to re-caption the images instead of relying on the original alt-text annotations, which are often noisy. However, we retain raw captions with high CLIP scores and randomly choose between VLM-generated captions and raw captions for these samples. We make this choice primarily to support keyword-based prompting such as ``high quality'' and ``4k'' during inference, since such keywords do not naturally emerge in VLM-generated captions.

\subsection{Training Setup and Hyperparameters}

We adopt the Emu-3.5 tokenizer \cite{cui2025emu3}, which has a vocabulary size of 131,072. We initialize \ours~from a pretrained diffusion language model \cite{fu2026nextron}. Training consists of two stages. In the first stage, we pretrain the model on $256 \times 256$ images for 200k steps with a global batch size of 1024. In the second stage, we scale to $512 \times 512$ resolution for 20k steps and then to $1024 \times 1024$ resolution for 80k additional steps with a global batch size of 256. Training is conducted on 64 H100 GPUs. Additional details are provided in Table~\ref{tab:training-stages}.

\begin{table}[t]
\centering
\caption{\textbf{Training configurations across two stages.}}
\label{tab:training-stages}
\begin{tabular}{lccH}
\toprule

 & \textbf{Stage 1} & \textbf{Stage 2} & \textbf{Stage 3} \\
\midrule
Learning Rate & $1 \times 10^{-4}$  & $1 \times 10^{-5}$ & $2 \times 10^{-5}$ \\
Steps & 200k & 100k & 100k \\
$\beta_1$ & 0.99 &0.99  & 0.99 \\
$\beta_2$ & 0.999 &0.999  & 0.999 \\
optimizer & AdamW & AdamW & AdamW \\
Learning Rate Schedule & Cosine  & Cosine & Cosine \\
Final  Learning Rate  & $1 \times 10^{-5}$  & $1 \times 10^{-6}$ & $1 \times 10^{-6}$ \\
\midrule
Model Size & 8B & 8B & 10.4B \\
Image Resolution & 256 & 512 $\rightarrow$ 1024 & 1024 \\
Global Batch Size & 1,024 & 256 \\
Token Editing & Disabled & Enabled \\
\bottomrule
\end{tabular}%

\end{table}

\subsection{Ablation Studies of Editing Thresholds}

\begin{figure}
    \centering
    \includegraphics[width=0.8\linewidth]{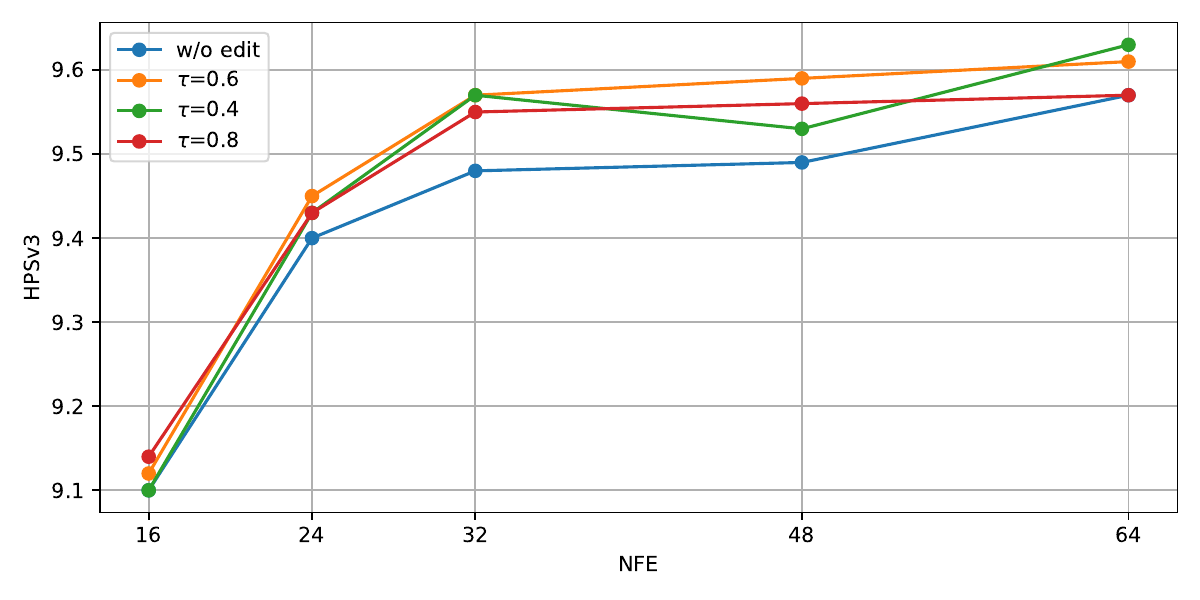}
    \caption{\textbf{Effect of different token-editing thresholds on HPSv3 scores.}}
    \label{fig:token_edit_thres}
\end{figure}

We study the effect of varying the editing threshold $\tau$ under different numbers of inference steps. We visualize the resulting HPSv3 scores in Figure~\ref{fig:token_edit_thres}. Overall, enabling token editing consistently outperforms the no-editing baseline. Among all evaluated settings, $\tau=0.6$ achieves the best image quality for most NFEs.

\subsection{Ablation Studies of Cluster Sizes}

We investigate the effect of varying the number of clusters used in the GCE objective and report FID scores on ImageNet-256 in Table~\ref{tab:snce_results_appendix}. We find that combining both 16,384-cluster and 8,192-cluster supervision performs better than using either clustering level alone. When using only a single clustering level, the 16,384-cluster setting performs better, presumably because the 8,192-cluster setting is coarser and provides less refined supervision signals.

\begin{table}[t]
\centering
\caption{\textbf{Ablation Experiments on Cluster size.}  *Model have identical-sized transformer layer. Parameter count increased due to larger token embedding and final linear head.}
 \label{tab:snce_results_appendix}
  \begin{adjustbox}{max width=\textwidth}
\begin{tabular}{lcccccc}
\toprule
\textbf{Cluster Sizes} & \textbf{Params} & \textbf{Tokenizer} &\textbf{Tokenizer Pretraining} &\textbf{Codebook} & \textbf{Epoch} & \textbf{FID} $\downarrow$ \\
\midrule
8,192  & 577M* &  Emu3.5-IBQ \cite{cui2025emu3} & Large-Scale T2I & 131,072 & 100 & 3.67 \\
16,384 & 577M* & Emu3.5-IBQ \cite{cui2025emu3}  & Large-Scale T2I  &131,072 & 100 & 3.44 \\
Both & 577M* & Emu3.5-IBQ \cite{cui2025emu3}  & Large-Scale T2I  &131,072 & 100 & 3.40 \\

\bottomrule
\end{tabular}
\end{adjustbox}
\end{table}

\subsection{Ablation Studies of Corruption Type and Scale}

We explore different corruption strategies for the token-editing objective by evaluating HPSv3 scores after 10K steps 1024 resolution training in Stage-2. Specifically, we evaluate random noise corruption, corruption using neighboring tokens in embedding space, and resampling tokens from the same input image. Results are reported in Table~\ref{tab:noise_thres}. Using a combination of neighboring tokens and resampled tokens achieves the best performance.

Additionally, we experiment with different corruption ratios $\alpha$ and observe no significant differences in image quality when $\alpha$ lies within a reasonable range. However, image quality degrades when $\alpha$ becomes too large. This is because high corruption levels (e.g., $\alpha=0.5$) make it substantially more difficult to distinguish clean tokens and corrupted tokens due to the lower signal-to-noise ratio, increasing optimization difficulty. In our final experiments, we use $\alpha=0.1$.

\begin{table}[h]
    \centering
    \begin{minipage}{0.45\textwidth}
        \centering
        \caption{Noise Type Comparison}
        \label{tab:noise_type}
        \begin{tabular}{lc}
            \hline
            \textbf{Noise Type} & \textbf{HPSv3 (10k step)} \\ \hline
            Random & 8.53 \\
            Adjacent Tokens & 8.81 \\
            Adj. Tokens + Resamp. & 8.99 \\ \hline
        \end{tabular}
    \end{minipage}
    \hfill 
    \begin{minipage}{0.45\textwidth}
        \centering
        \caption{Threshold $\alpha$ Impact}
                \label{tab:noise_thres}
        \begin{tabular}{lc}
            \hline
            \textbf{$\alpha$ Value} & \textbf{HPSv3 (10k step)} \\ \hline
            0.1 & 8.99 \\
            0.3 & 8.97 \\
            0.5 & 7.52 \\ \hline
        \end{tabular}
    \end{minipage}
\end{table}

\subsection{Additional Qualitative Results}

\begin{figure}
    \centering
    \includegraphics[width=1.0\linewidth]{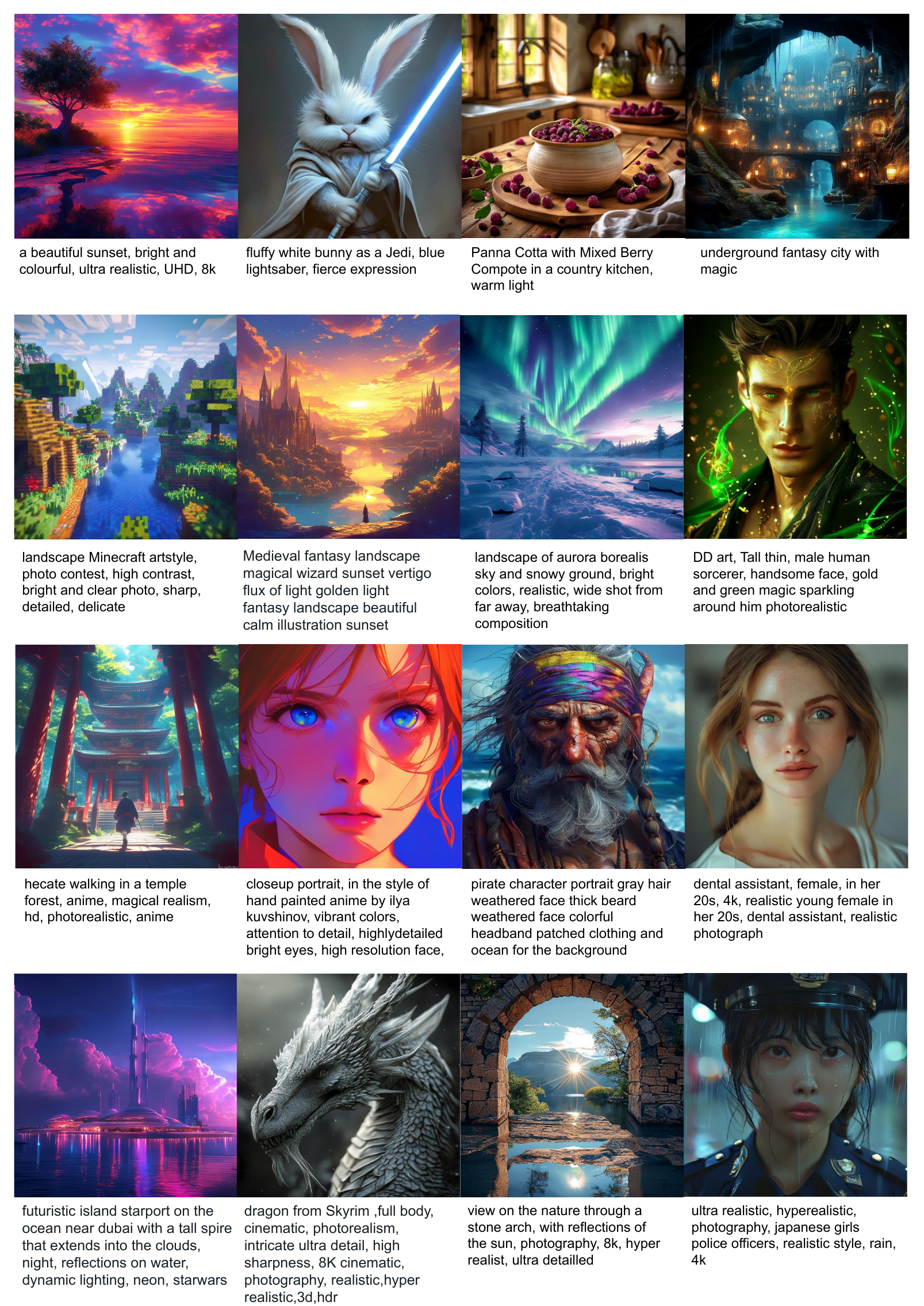}
    \caption{Additional qualitative text-to-image generation results.}
    \label{fig:appendx_demo}
\end{figure}

\begin{figure}
    \centering
    \includegraphics[width=0.9\linewidth]{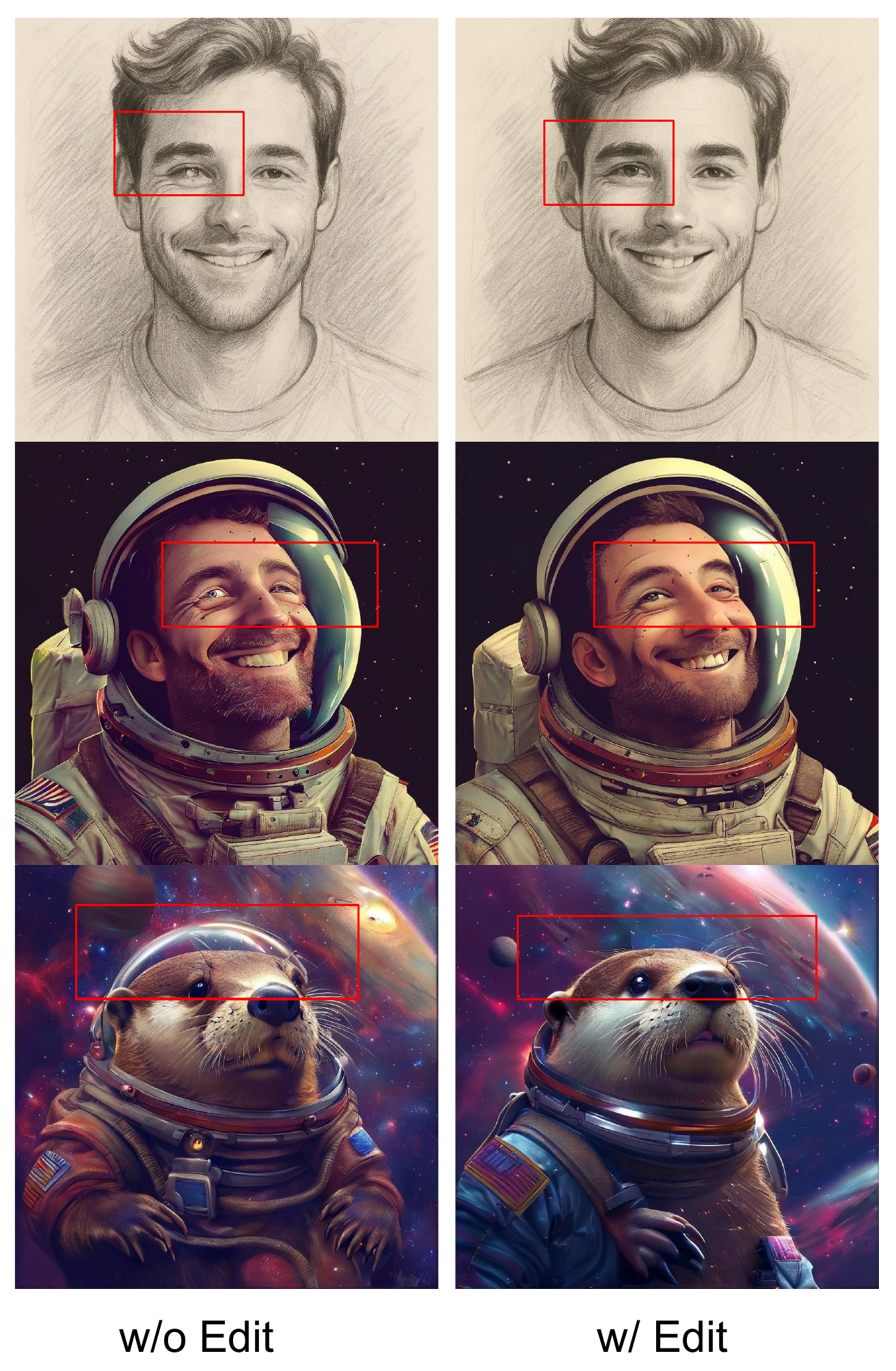}
    \caption{Qualitative comparison between generations with and without token editing.}
    \label{fig:appendx_demo_edit}
\end{figure}

\begin{figure}
    \centering
    \includegraphics[width=0.9\linewidth]{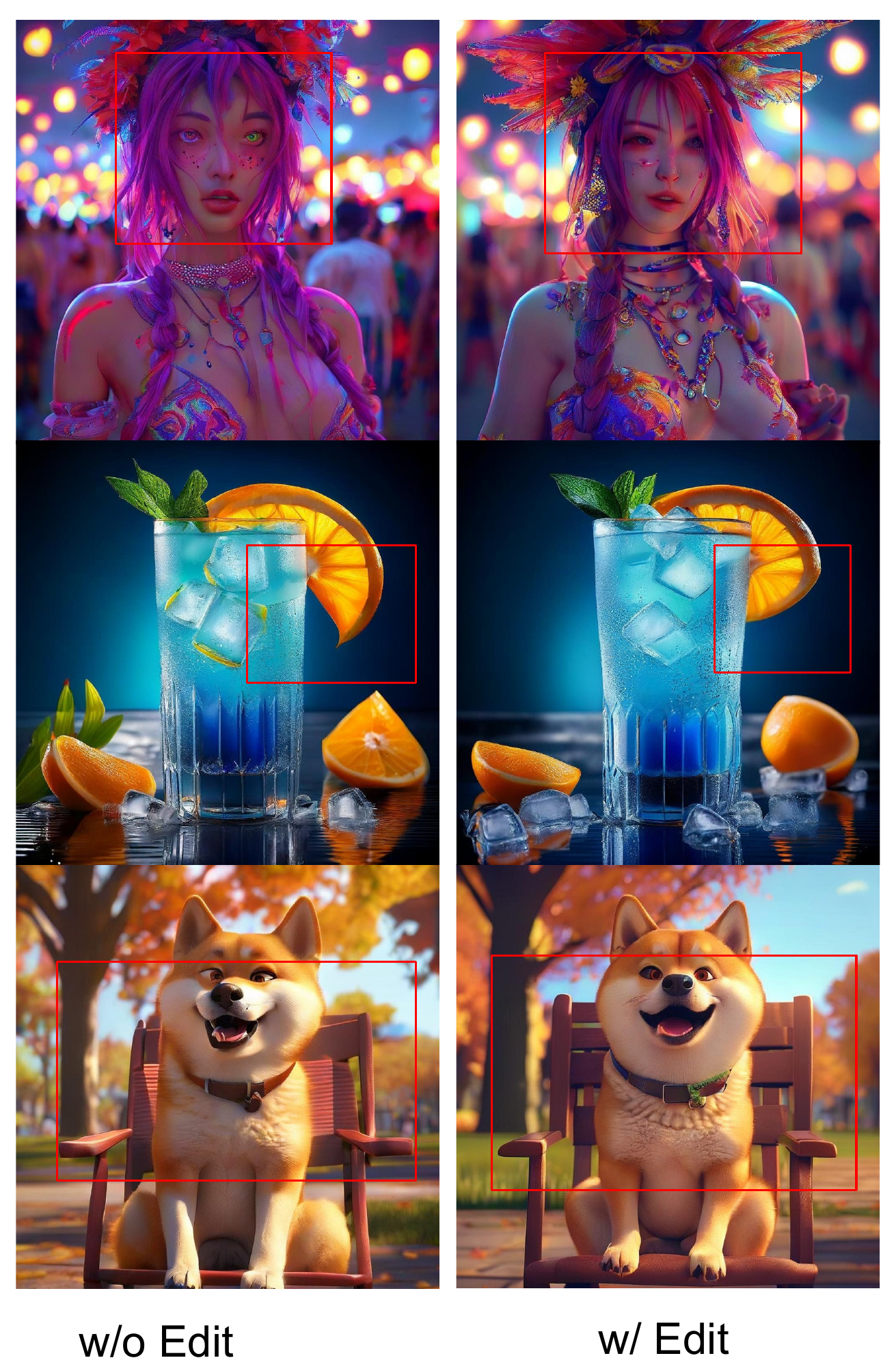}
    \caption{Additional qualitative comparisons for token editing.}
    \label{fig:appendx_demo_edit2}
\end{figure}

In this section, we provide additional qualitative samples to further demonstrate the effectiveness of \ours. Figure~\ref{fig:appendx_demo} presents additional text-to-image generation results. Figures~\ref{fig:appendx_demo_edit} and \ref{fig:appendx_demo_edit2} compare generations produced with and without token editing under the same random seed. We observe that token editing consistently improves image fidelity by refining details and correcting artifacts.

\section{Compute Resources}
\label{sec:compute}

We train the model on 64 H100 GPUs across 8 nodes. Training takes approximately 16 days in total.

\section{Limitations}
\label{sec:limitation}

Despite the effectiveness of \ours, several limitations remain. First, although we demonstrate that the token-editing mechanism improves image quality, it does not eliminate all artifacts, and the model may still generate erroneous outputs. Second, although we achieve substantial performance gains by optimizing the GCE objective with a carefully designed fused operator, additional improvements may still be possible through customized low-level CUDA kernels. We leave this direction for future work.

\section{Broader Impact}
\label{sec:boarder_impact}

\ours~has strong text-to-image generation capabilities, which may be misused to generate harmful or offensive content. We strongly caution against such use cases. Additionally, our model may inherit biases present in the base language model as well as biases contained in the training data. Our model is intended primarily for research purposes to facilitate future exploration of foundational discrete image generators. We do not recommend its use for other purposes.

\section{Licenses}
\label{sec:Licenses}

We make use of the following assets:

\textbf{Models:} Emu-3.5-Tokenizer \cite{cui2025emu3} (Apache-2.0), Qwen3-VL \cite{bai2025qwen3} (Apache-2.0), Nemotron-Labs-Diffusion \cite{fu2026nextron} (Nvidia Open Model License)

\textbf{Datasets:} LAION \cite{schuhmann2022laion} (MIT), COYO \cite{kakaobrain2022coyo-700m} (CC-BY-4.0), MJHQ \cite{li2024playground} (CC-BY-4.0), BLIP3o-60k \cite{chen2025blip3} (Apache-2.0), and ShareGPT4o-Image \cite{chen2025sharegpt} (CC-BY-4.0).

\end{document}